\documentclass[review]{elsarticle}
\usepackage{subfigure}
\usepackage{geometry}
\usepackage{color}
\usepackage{amsmath} % assumes amsmath package installed
\usepackage{amssymb}  % assumes amsmath package installed
\usepackage{graphicx}
\usepackage{psfrag}
\usepackage{epstopdf}
\usepackage{epsfig,graphics,subfigure,psfrag,amsmath,amssymb,diagbox}
\usepackage{float}
\usepackage{subfigure}

\usepackage{lineno,hyperref}
\usepackage{multirow}
\modulolinenumbers[5]
\usepackage{array}
\usepackage{booktabs}
\usepackage{longtable}
\usepackage{makecell}
\usepackage{caption}
\usepackage{chngpage}

\usepackage{bm}
\usepackage{colortbl}
\usepackage{arydshln}
\usepackage{multicol}
\usepackage{longtable}
\usepackage{multirow} % added by Feng Zhou
\usepackage{titlesec}
\titleformat{\section}[hang]{\large\bfseries\sffamily}{\thesection.}{0.3em}{}
\titleformat{\subsection}[hang]{\bfseries\sffamily}{\thesubsection.}{0.3em}{}
\usepackage[aboveskip=1pt,labelfont=bf,labelsep=period]{caption}

\usepackage{diagbox}
\usepackage{framed}
\usepackage{soul}
\usepackage{ulem}

\usepackage{picins}

\usepackage[ruled, linesnumbered]{algorithm2e}

\newtheorem{thm}{Theorem}[section]

\newtheorem{defn}{Definition}[section]

\newtheorem{rem}{Remark}
\newproof{pf}{Proof}

\biboptions{numbers,sort&compress}

\usepackage{setspace}

\journal{Pattern Recognition}

\begin{document}
\begin{frontmatter}

\title{RRCNN: A novel signal decomposition approach based on recurrent residue convolutional neural network}

\author[gdufe]{Feng Zhou\corref{cor}}         
\address[gdufe]{School of Information Sciences, Guangdong University of Finance and Economics, Guangzhou, 510320, China}
\ead{fengzhou@gdufe.edu.cn}
\cortext[cor]{Corresponding author}

\author[antonio1,antonio2,antonio3] {Antonio Cicone}
\address[antonio1]{Department of Information Engineering, Computer Science and Mathematics, University of L'Aquila, L'Aquila, 67100, Italy}
\address[antonio2]{Istituto di Astrofisica e Planetologia Spaziali, INAF, Rome, 00133, Italy}
\address[antonio3]{Istituto Nazionale di Geofisica e Vulcanologia, Rome, 00143, Italy}
\ead{antonio.cicone@univaq.it}

\author[gatech]{Haomin Zhou}
\address[gatech]{School of Mathematics, Georgia Institute of Technology, Atlanta, GA 30332, United States}
\ead{hmzhou@math.gatech.edu}

\begin{abstract}
The decomposition of non-stationary signals is an important and challenging task in the field of signal time-frequency analysis. In the recent two decades, many signal decomposition methods led by the empirical mode decomposition, which was pioneered by Huang et al. in 1998, have been proposed by different research groups. However, they still have some limitations. For example, they are generally prone to boundary and mode mixing effects and are not very robust to noise. Inspired by the successful applications of deep learning in fields like image processing and natural language processing,  and given the lack in the literature of works in which deep learning techniques are used directly to decompose non-stationary signals into simple oscillatory components, we use the convolutional neural network, residual structure and nonlinear activation function to compute in an innovative way the local average of the signal, and study a new non-stationary signal decomposition method under the framework of deep learning. We discuss the training process of the proposed model and study the convergence analysis of the learning algorithm. In the experiments, we evaluate the performance of the proposed model from two points of view: the calculation of the local average and the signal decomposition. Furthermore, we study the mode mixing, noise interference, and orthogonality properties of the decomposed components produced by the proposed method. All results show that the proposed model allows for better handling boundary effect, mode mixing effect, robustness, and the orthogonality of the decomposed components than existing methods.
\end{abstract}

\begin{keyword}
Empirical mode decomposition; Adaptive signal decomposition; Signal local average; Convolutional neural network; Residual network
\end{keyword}

\end{frontmatter}

%\linenumbers

\vspace{0.5em}
\section{Introduction}
\label{sec::introduction}
With the development of technology, many every day--signals that exhibit nonlinearity and non-stationarity, such as human speech, radar systems, and seismic waves, can be accurately captured. It is well known that decomposing and exploring features of this kind of signals is quite challenging due to their nonlinear and non-stationary characteristics.

In the past two decades, many studies have emerged for processing non-stationary signals. One of the most representative works is the empirical mode decomposition (EMD) algorithm along with the Hilbert spectrum analysis proposed by Huang et al. in 1998 \cite{huang1998empirical}. Because EMD is fully data-driven, and can adaptively decompose a signal into several intrinsic mode functions (IMFs), it has already shown its usefulness in a wide range of applications, including semantic recognition \cite{aksoy2017semantic}, alcoholism identification \cite{thilagaraj2019empirical}, and stock trend prediction \cite{zhou2019emd2fnn}. Despite its remarkable success, it still lacks mathematical foundations and is sensitive to noise and sampling. This sparked many efforts to improve the EMD. The improvements share the same feature: a signal is decomposed into several simpler components, and then a time-frequency analysis method is applied to each component separately. These signal decomposition methods can be mainly achieved in two ways: by iteration or by optimization.

Methods based on iteration include many techniques, such as moving average, partial differential equation (PDE) and filter. For instance, Smith presented a new iteration method, based on the local average, to decompose the non-stationary signals into a set of functions \cite{smith2005local}. Del{\'e}chelle et al. proposed a new approach that resolves one major problem in the EMD, that is, the mean envelope detection of a signal, in virtue of a parabolic PDE \cite{delechelle2005empirical}. Hadji et al. used the differential calculus on envelopes, which makes them prove that iterations of the sifting process are well approximated by the resolution of PDE \cite{el2009analysis}. Hong et al. introduced a novel sifting method based on the concept of the local integral mean of a signal \cite{hong2009local}. And Cicone et al. studied the method based on iterative filtering to compute the local average, which is utilized to replace the mean of the upper and lower envelopes in the sifting procedure of the EMD \cite{cicone2016adaptive, cicone2022multivariate}. Tu et al. proposed the iterative nonlinear chirp mode decomposition (INCMD) \cite{tu2020iterative} under the framework of the variational nonlinear chirp mode decomposition.

On the other hand, there are methods based on optimization. Peng et al. designed an adaptive local linear operator-based optimization model to decompose a signal into several local narrow band signals \cite{peng2010null}. Oberlin et al. proposed an optimization model in computing the mean envelope to replace the original one in EMD \cite{oberlin2012alternative}. Inspired by the compressed sensing theory, Hou et al. studied a new adaptive data analysis method, which can be seen as a nonlinear version of compressed sensing and provides a mathematical foundation of the EMD method \cite{hou2011adaptive}. Flandrin et al. proposed a convex optimization procedure in order to replace the sifting process in the EMD, which follows the idea of texture-geometry decomposition with further specific EMD features such as quasi-orthogonality and extrema-based constraints \cite{pustelnik2012multicomponent, pustelnik2014empirical}. Dragomiretskiy et al. put forward the variational mode decomposition (VMD), whose goal is to decompose a signal into a discrete number of modes, that have specific sparsity properties while reproducing the input \cite{dragomiretskiy2013variational}. Rehman et al. generalized the VMD method to multivariate or multichannel data \cite{ur2019multivariate}. And Zhou et al. presented a new mathematical framework by finding the local average based on the local variational optimization model \cite{zhou2016optimal}.

In addition, there are some methods that cannot be classified into the above two categories. For instance, Daubechies et al. proposed the method, called synchrosqueezed wavelet transforms, by combining the wavelet analysis and reallocation method \cite{daubechies2011synchrosqueezed}. Gille presented the approach, called empirical wavelet transform (EWT), to build adaptive wavelets \cite{gilles2013empirical}, whose main idea is to extract the different modes by designing an appropriate wavelet filter bank. Singh et al. studied the adaptive Fourier decomposition method (FDM) based on the Fourier theory, which decomposes any data into a small number of ``Fourier intrinsic band functions" \cite{singh2017fourier, singh2018novel}. And Wang et. extended the adaptive FDM to the multi-channel case \cite{wang2022adaptive}.

According to the works described above, we find that whether the method is based on iterative or optimization, calculating the local average of a given signal is very critical. For example, in EMD \cite{huang1998empirical}, the mean of the upper and lower envelopes are used to measure the local average of the signal; the local variational optimization model is constructed to compute the local average in \cite{zhou2016optimal}; and in the iterative filtering method \cite{cicone2016adaptive,cicone2022multivariate}, the low-pass filter is employed to find the local average. Although there exist many studies on the characterization of the local average, it is basically impossible to find a method suitable for all signals from a practical point of view. Discussing the local average customized according to the type of signal, it not only provides a new research perspective, but also is likely to become the trend in the near future in signal processing for non-stationary data.

In recent years, thanks to the remarkable results obtained in fields of research like image and natural language processing, the usage and application of deep learning methods have spread widely in an ample variety of research fields, like image processing \cite{hemanth2017deep} and natural language processing \cite{li2017deep}. In signal processing, deep learning models have been used, so far, to achieve various goals, such as: noise removal \cite{rudy2019deep,zhu2019seismic}, forecasting \cite{zheng2020ensemble,sahoo2022deep, ding2022mine}, and detection \cite{barz2018detecting, zhang2021learning}. However, to the best of our knowledge, not a single method has been proposed so far in the literature, which allows to decompose a given non-stationary signal into simple oscillatory components, like the IMFs, which is solely based on deep learning techniques.

For this reason, in the current work we propose an innovative signal decomposition algorithm, named recurrent residual convolutional neural network (RRCNN), which is based on deep learning models. In the RRCNN method, in fact, we first characterize the local average of a signal under the framework of deep learning, and then use it to handle signal decomposition. Specifically, the 1-Dimensional (1-D) convolutional neural network is primarily designed to adaptively compute the local average of the input signal, which is similar to the moving average method and the filter operation in the iterative filtering method, except that the weights in 1-D convolutional neural network are not fixed, but are learned adaptively during the training phase according to the input signals. Moreover, both the residual and recurrent structures are employed to amend the computed local average, which is consistent with the role of the inner loop process in many existing iterative-based signal decomposition methods. After the local average model is derived, it is cascaded in series to realize the decomposition model of the signal, whose function is equivalent to the outer loop structure of the existing iterative-based signal decomposition methods.

Although the proposed method looks similar to those iterative-based decomposition methods that contain a two-loop structure, the use of the deep learning techniques makes this method have the following peculiarities:
\begin{itemize}
\item[(\romannumeral1)] Unlike the moving average method and the filter operation in the iterative filtering method, the convolutional filter weights that appear in the proposed RRCNN model, are not fixed in advance, but are learnt adaptively in the training phase according to the inputs.
\item[(\romannumeral2)] Since the proposed RRCNN model is constructed under the framework of deep learning, it makes RRCNN more flexible and adaptive in finding the local average and achieving the decomposition for a given signal. In particular, the nonlinear activation function can be added after the convolutional operation to increase the expression ability. The loss function also can be customized according to the requirements that the ground truths usually have in the specific application. \item[(\romannumeral3)] Several artificial signals are constructed to verify the performance RRCNN in terms of local average characterization, noise interference, mode mixing, and orthogonality. Furthermore, we compare the RRCNN model with the state-of-the-art methods. In addition, we also use the solution of the Duffing and Lorenz equations, and the real data of the length of day (LOD) to evaluate the approximation ability of RRCNN to the existing models.
\item[(\romannumeral4)] Generally speaking, the RRCNN model takes a certain amount of time in the training phase, which is a commonality of deep learning-based models. However, once the model training is completed, the computational efficiency in the prediction stage is relatively fast, especially it can use the parallelization mechanism to predict multiple signals at the same time, which is not available in most existing methods.
\item[(\romannumeral5)] RRCNN has the limitations brought from the supervised model. For example: in the training phase, each input signal needs to know its label in advance.
\end{itemize}

The rest of the paper is organized as follows. We review the iterative filtering method and provide its algorithm in Section \ref{sec:IF}. And the concept of $\beta$-smooth function and its properties are given in Section \ref{sec::beta_smooth}, which are used for proving the convergence of the proposed model. In Section \ref{sec:RRCNN,Multi-RRCNN}, the new local average method and the derived signal decomposition method, collectively called the RRCNN, are proposed. Moreover, the training process and convergence analysis of RRCNN are given in this section. In Section \ref{sec:experiments}, we study a series of examples to evaluate the performance of RRCNN compared with the existing methods. Finally, we give the conclusion in Section \ref{sec:conclusion}.

\section{IF and $\beta$-smooth function}
\subsection{IF}
\label{sec:IF}

The iterative filtering (IF) \cite{cicone2016adaptive} is a recurrent algorithm that decomposes a nonlinear and non-stationary signal into a number of IMFs. The main idea of IF is the subtraction of local moving averages from the signal iteratively, where the local moving averages are calculated through convolutions with low-pass filters. Alg. \ref{alg::IF} shows the detailed steps, where the parameter $l_n$, called the filter length, is important in the IF method, and is determined by the information contained in the signal itself; {$w_n(\cdot)$} represents the low-pass filter function.

\begin{algorithm}[hbt!]
\caption{Iterative filtering (IF)}\label{alg::IF}
\KwData{Given a signal $x(t)$}
\KwResult{$IMF$}
%$y \gets 1$\;
%$X \gets x$\;
%$N \gets n$\;
\While{the number of extrema of $x\geq 2$}{
  %\eIf{$N$ is even}{
    $n=0$\; 
    $x_1(t)=x(t)$\;
    \While{the stopping criterion is not satisfied}{
     compute the filter length $l_n$ and filter weight function $w_n$ for $x_n$\;
     $x_{n+1}(t)=x_{n}(t)-\int_{-l_{n}}^{l_{n}}x_{n}(t+y)w_n(y)dy$\;
     $n=n+1$\;
    }
   $IMF=IMF\cup\{x_{n}\}$\; % \Comment*[r]{This is a comment}
%  }{\If{$N$ is odd}{
      $x(t)=x(t)-x_{n}(t)$\;
    %}
 % }
}
$IMF=IMF\cup \{x\}$.
\end{algorithm}

\subsection{$\beta$-smooth function and some of its properties}
\label{sec::beta_smooth}

We first introduce the concepts of $L$-Lipschitz continuous and $\beta$-smooth for a function from \cite{bubeck2015convex}.

\begin{defn}A function $f$ is said to be $L$-Lipschitz continuous if for all $x,y\in\mathcal{X}$, $\|f(x)-f(y)\|\leq L\|x-y\|$, where $\mathcal{X}$ denotes the convex domain of $f$, and $L$ is called the Lipschitz constant.
\end{defn}

\begin{defn}\label{def::beta_smooth}
A continuously differentiable function $f$ is $\beta$-smooth if the gradient $\nabla f$ is $\beta$-Lipschitz, that is if for all $x,y\in\mathcal{X}$, $\| \nabla f(x)-\nabla f(y)\|\leq \beta\|x-y\|$, where $\mathcal{X}$ is the convex domain of $f$.
\end{defn}

Then, for a unconstraint optimization problem, if its objective function is $\beta$-smooth, we can prove that the sequence generated by the gradient descent algorithm converges to a stationary point when the learning rate is small enough. The details can be found in Theorem \ref{thm::convergence}.

\begin{thm}\label{thm::convergence}
Let $f$ be a $\beta$-smooth function and $f^{*}=\min f(x)>-\infty$. Then the gradient descent algorithm with a constant learning rate $\lambda<\frac{2}{\beta}$, i.e.,
$x^{(k+1)}=x^{(k)}-\lambda \nabla f(x^{(k)}),$ converges to a stationary point, i.e., the set $\{x:\nabla f(x)={\bf0}\}$.
\end{thm}

\begin{pf}
According to the gradient descent algorithm, i,e.,
\begin{equation}
\label{equ::gd_proof}
x^{(k+1)}=x^{(k)}-\lambda\nabla f(x^{(k)}),
 \end{equation}
 as $f$ is $\beta$-smooth, we have
\begin{equation}\nonumber\begin{split}
f(x^{(k+1)})&\overset{(a)}{\leq} f(x^{(k)})+\nabla f(x^{(k)})(x^{(k+1)}-x^{(k)})+\frac{\beta}{2}\|x^{(k+1)}-x^{(k)}\|^2\\
&\overset{(b)}{=} f(x^{(k)})-\lambda\|\nabla f(x^{(k)})\|^2+\frac{\beta \lambda^2}{2}\|\nabla f(x^{(k)})\|^2\\
&=f(x^{(k)})-\lambda(1-\frac{\beta \lambda}{2})\|\nabla f(x^{(k)})\|^2,
\end{split}\end{equation}
where the inequality (a) follows from Lemma 3.4 in \cite{bubeck2015convex}, and the equality (b) is obtained from Eqn. (\ref{equ::gd_proof}). Due to $\lambda<2/\beta$, it becomes
\begin{equation}\nonumber
\|\nabla f(x^{(k)})\|^2\leq \frac{f(x^{(k)})-f(x^{(k+1)})}{\lambda(1-\frac{\beta\lambda}{2})}.
\end{equation}
Next, we have
\begin{equation}\nonumber
\sum_{k=0}^{K}\|\nabla f(x^{(k)})\|^2\leq \frac{1}{\lambda(1-\frac{\beta\lambda}{2})}\sum_{k=0}^{K}(f(x^{(k)})-f(x^{(k+1)}))=\frac{f(x^{(0)})-f(x^{(K+1))}}{\lambda(1-\frac{\beta\lambda}{2})} \leq \frac{f(x^{(0)})-f(x^*)}{\lambda(1-\frac{\beta\lambda}{2})},
\end{equation}
where $x^{*}$ denotes the global optimization point. Taking the limit as $K\rightarrow +\infty$, we have
%\begin{equation}\nonumber
$\sum_{k=0}^{+\infty}\|\nabla f(x^{(k)})\|^2\leq +\infty.$
%\end{equation}
Hence, $\lim_{k\rightarrow +\infty}\nabla f(x^{(k)})=0$ is obtained.
\end{pf}

\section{RRCNN inner loop block and RRCNN}
\label{sec:RRCNN,Multi-RRCNN}

\subsection{RRCNN inner loop block}
\begin{figure}[H]
\centering
\includegraphics[width=4.5in]{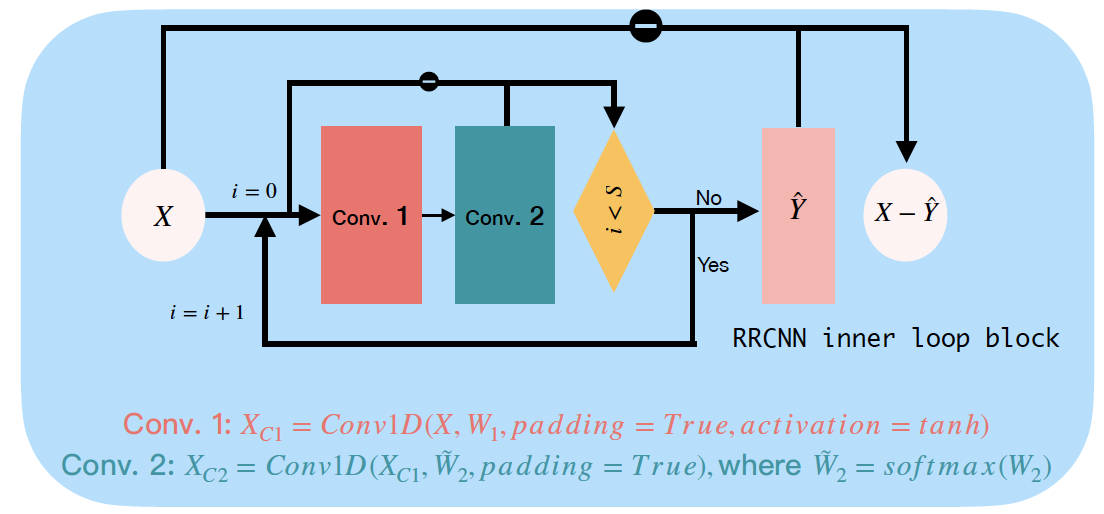}
\caption{Graphic illustration of the RRCNN inner loop block.}
\label{fig:ResBlock}
\end{figure}

The main operation in IF is the computation of moving average, which is essentially realized by the convolution operation, where the filter length depends on the given signal, and the filter weights are mainly given by some empirical functions selected artificially a priori. Therefore, it is very natural to convert the convolution operation into a 1-D convolutional neural network model, where both the filter length and the filter weights can be learnt adaptively according to the input signals given in advance. Furthermore, some ingenious mechanisms in deep learning, such as the nonlinear activation function, the residue learning \cite{he2016deep}, etc., can be adopted to make it more flexible. The structure we design to mimic the inner while loop of Alg. \ref{alg::IF}, is graphically depicted in Fig. \ref{fig:ResBlock}. Since it mainly contains the recurrence mechanism, the convolutional layer and the subtraction operation, we call it the recurrent residual convolutional neural network (RRCNN) inner loop block.

\begin{algorithm}
\caption{RRCNN inner loop block} \label{alg::RRCNN}
\KwData{$X\in\mathbb{R}^{N}$}
\KwResult{the local average of $X$}
 Initialize $i=0$ and $X^{(0)}=X$\;
 \While{$i<S$}{
  The input $X^{(i)}$ goes through the first 1-D convolutional layer, i.e., $X_{C1}:=Conv1D(X^{(i)}, W_1^{(i)}, padd$- $ing=True, activation=\tanh)$\;
  Transfer $X_{C1}$ to the second convolutional layer, i.e., $X_{C2}:=Conv1D$ $(X_{C1}, \tilde{W}_2^{(i)}, padding=True)$, where $\tilde{W}_2^{(i)}:=softmax(W_2^{(i)})$\;
   $X^{(i+1)}=X^{(i)}-X_{C2}$\;
   $i=i+1$\;
 }
  $\hat{Y}=X^{(S)}$ and $X-\hat{Y}$ are the IMF and the local average respectively.
\end{algorithm}

As shown in Fig. \ref{fig:ResBlock}, the inner loop mainly consists of a judgment-loop structure and a residue operation, and the judgment-loop structure is formed of two convolutional layers and a residual operation. Suppose $X\in\mathbb{R}^{N}$ {(the vectors in this article are column vectors by default unless otherwise specified.)} denote the input, the output of the RRCNN inner loop block, called {$\hat{Y}\in\mathbb{R}^N$}, is computed as the following Alg. \ref{alg::RRCNN}. {And} {$X-\hat{Y}$} is the local average of the input signal obtained by the RRCNN inner loop block.

Mathematically, the output of the RRCNN inner loop block can be expressed as:
$\hat{Y}=F(X, {\bf W})$, where $\bf W$ denotes the undetermined weights in the RRCNN inner loop block, and the function $F$ represents the structure of RRCNN inner loop block, which is composed of the operators including convolution, nonlinear activation and subtraction. The detailed process of $F$ can be formulated as:
$F(X,{\bf W})=f(X^{(S-1)},{\bf W}^{(S-1)})$, where $S$ represents the number of recursion in the RRCNN inner loop block, ${\bf W}^{(S-1)}$ is the undetermined weights in the $(S-1)$-th recursion, and the function $f$ and $X^{(S-1)}$ are defined as:
\begin{equation}\left\{\begin{split}
%\label{equ::RRCNN_block_f}
&f(X^{(i)},{\bf W}^{(i)})=\tanh(X^{(i)}\ast W_{1}^{(i)})\ast \tilde{W}_{2}^{(i)},\\
&X^{(i+1)}=X^{(i)}-f(X^{(i)},{\bf W}^{(i)}),
\end{split}\right.\end{equation}
where $X^{(0)}=X$, ${\bf W}^{(i)}$ is the undetermined weights in the $i$-th recursion that it includes the weights, denoted as $W_1^{(i)}\in\mathbb{R}^{K_1}$ and $W_2^{(i)}\in\mathbb{R}^{K_2}$, $\tilde{W}_2^{(i)}:=softmax(W_2^{(i)})=\{\frac{exp({{W_2^{(i)}}_l})}{\sum_k exp({{W_2^{(i)}}_k})}\}_{l=1}^{K_2}$, $\ast$ is the 1-D convolution operation, and $i=0, 1, \ldots, S-2$.

It is worth pointing out that the roles of the two 1-D convolutional layers in each judgment-loop are different. The role of the first convolutional layer, which is configured with a non-linear activation function (we select $\tanh$ in this work), is to enhance the nonlinear expression ability of the method. Whereas, the purpose of the second convolutional layer is to make the result more reasonable to describe the local average of the signal. Therefore, the non-negativity and normalization restrictions of its weights are added; and there is no nonlinear activation function configured with it. The use of padding in the two layers is to ensure that the length of the output is consistent with the input.
We will discuss the training details of the RRCNN inner loop block in the following section.

\subsection{RRCNN}\label{subsec::RRCNN}

\begin{figure}[H]
\centering
\includegraphics[width=6.8in]{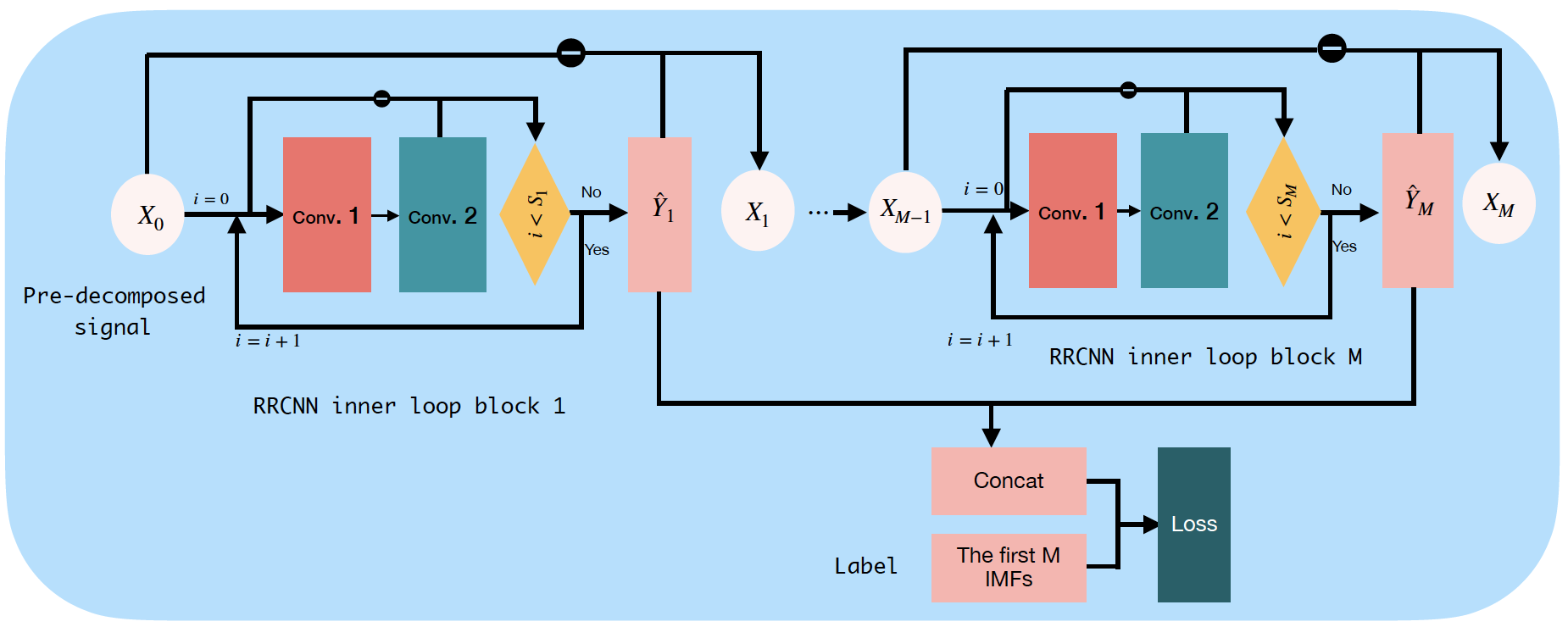}
\caption{Graphic illustration of the RRCNN.}
\label{fig:CNN}
\end{figure}

After the RRCNN inner loop block for the identification of the local average of a given signal is constructed, we can cascade a finite number of RRCNN inner loop blocks together to derive the signal decomposition, which is called RRCNN, and is shown in Fig. \ref{fig:CNN}. According to it, an input signal $X\in\mathbb{R}^{N}$ (also denoted as $X_0$) can be decomposed into $M$ IMFs {$\hat{\bf Y}:=\{\hat{Y}_i\}_{m=1}^{M}$ (each $\hat{Y}_i\in\mathbb{R}^N$) and a residue $X_M\in\mathbb{R}^{N}$}. The detailed steps of RRCNN are listed in Alg. \ref{alg::multi-RRCNN}.

The output of RRCNN can be formulated as:
\begin{equation}
\left\{\begin{split}
\label{equ::RRCNN}
&\hat{Y}_m=F(X_{m-1}, {\bf W}_m), \\
&X_m=X_{m-1}-\hat{Y}_m,
\end{split} \right.
\end{equation}
where $m=1,2,\ldots,M$, $X_0=X$, $F(X_{m-1}, {\bf W}_m)$ is the $m$-th RRCNN inner loop block whose purpose is to extract an IMF from $X_{m-1}$, and ${\bf W}_m$ denotes the undetermined weights of the $m$-th RRCNN inner loop block.

\begin{algorithm}
\caption{RRCNN} \label{alg::multi-RRCNN}
\KwData{$X\in\mathbb{R}^{N}$, and the number of IMFs $M$}
\KwResult{the IMFs and residue of $X$}
 Initialize $m=1$, $X_0=X$\;
 \While{$m\leq M$}{
 	Compute the $m$-th IMF and the local average for the input $X_{m-1}$, denoted as $\hat{Y}_{m}$ and $X_{m}$ respectively, according to the RRCNN inner loop block\;
	$m=m+1$\;
 }
 $\hat{\bf Y}=\{\hat{Y}_m\}_{m=1}^{M}$ and $X_{M}$ are the resulting IMFs and residue of RRCNN.
\end{algorithm}

All the generated IMFs are concatenated as the outputs of the RRCNN model. The errors between the outputs, i.e., $\hat{\bf Y}$, and the labels that are composed of the true first $M$ IMFs, denoted as ${\bf Y}\in\mathbb{R}^{N\times M}$, are computed by the loss function. For example, the loss function can be expressed as:
\begin{equation}
L(\hat{\bf Y}, {\bf Y})=\|\hat{\bf Y}-{\bf Y}\|_{F}^2,
\label{equ::loss_function}
\end{equation}
where the errors are measured by mean square error (MSE), and $\|\cdot\|_{F}$ denotes the Frobenius norm. In the RRCNN model equipped with the loss function as in Eqn. (\ref{equ::loss_function}), the computational complexity of the forward process of RRCNN is mainly attributed to the computation of the convolutional layer, which is $O(N\cdot K)$, where $N$ and $K$ denote the length of the signal and the size of the convolutional filter, respectively.

The loss can be customized according to the characteristic of the decomposition task. For example, if the 3rd IMF are smooth, the quadratic total variation term, expressed as $$QTV(\hat{\bf Y}_{\Omega_1}):=\sum_{m\in\Omega_1}\sum_{t=1}^{N-1}(\hat{Y}_{(t+1),m}-\hat{Y}_{t,m})^2,$$ can be added to the loss function, where $\Omega_1$ represents the set of subscripts of those smooth components (here $\Omega_1=\{3\})$, i.e.,
\begin{equation}
\label{equ::loss_function_TV}
L(\hat{\bf Y}, {\bf Y})=\|\hat{\bf Y}-{\bf Y}\|_{F}^2+\eta QTV(\hat{\bf Y}_{\Omega_1}),
\end{equation}
where $\eta\geq 0$ is a penalty parameter, $\hat{\bf Y}$ is the dependent variable of the function $F(\cdot, \cdot)$, and its independent variables are $X$ and $\bf W$ respectively.

Moreover, if the 2nd and 3rd IMFs are orthogonal, an orthogonal constraint can be added to the loss function to ensure the orthogonality of the resulting components, i.e.,
\begin{equation}\begin{split}
L(\hat{\bf Y}, {\bf Y})&=\sum_{i\in\hat{\Omega}_2^c}\|\hat{Y}_{i}-Y_{i}\|_{2}^2+\sum_{(i,j)\in\Omega_2} \|{\bf W}^{o_{ij}}{\hat{Y}_{i}}-{Y_{i}}\|_{2}^2+\|{\bf W}^{o_{ij}}{\hat{Y}_{j}}-{Y_{j}}\|_{2}^2,\\
& \text{ s.t. } {\bf W}^{o_{ij}}{{\bf W}^{o_{ij}}}^{\top}={\bf I},
\label{equ::loss_function_customized}
\end{split}\end{equation}
where ${\bf W}^{o_{ij}}\in\mathbb{R}^{N\times N}$ stands for the orthogonal matrix to be determined by $\min_{\{{\bf W}, {\bf W}^{o_{ij}}\}}L(\hat{Y}, Y),$ $\Omega_2$ (here $\Omega_2=\{(2,3)\}$) denotes the subscript pairs of those orthogonal components, $\hat{\Omega}_2$ (here $\hat{\Omega}_2=\{2,3\}$) represents the set consisting of all subscripts that appear in $\Omega_2$, and $\hat{\Omega}_2^c=\{1,2,\ldots,M\}-\hat{\Omega}_2$. In specific execution process, the orthogonal transformation ${\bf W}^{o_{ij}}$ of $\hat{Y}_i$ and $\hat{Y}_j$ can be regarded as adding a fully connected layer after the outputs of $\hat{Y}_i$ and $\hat{Y}_j$. The two fully connected layers share weights, i.e., ${\bf W}^{o_{ij}}$, and satisfy orthogonality, i.e., ${\bf W}^{o_{ij}}{{\bf W}^{o_{ij}}}^{\top}={\bf I}$. In this case, the result of any IMF whose subscript meeting $i\in\hat{\Omega}_2$ is updated from $\hat{Y}_i$ to ${{\bf W}^{o_{ij}}}\hat{Y}_i$, and the results of other components remain unchanged.

Compared with IF, RRCNN also contains two loops. The outer loop successively acquires $M$ IMFs and one residue. And the purpose of the inner loop, i.e., the RRCNN inner loop block, is mainly to compute the local average by several iterations, and finally get the IMF and local average of the current input signal. On the other hand, these two methods have important differences, which can be summed up as:
\begin{itemize}
\item[(\romannumeral1)] The number of outer loop of the IF method is data-driven, which might be different for different signals. While it is given in advance for the proposed RRCNN approach. Therefore, RRCNN is suitable for decomposing the non-stationary signals containing the same number of mono-components.
\item[(\romannumeral2)]  In the IF method, the filter length that is a key parameter, is determined only by the signal under decomposition. However, its filter weights lack of adaptability, and they are basically determined by the filter length. For the RRCNN approach, the filter length in each convolutional layer is adaptive, and can be selected by hyper-parameter optimization. Moreover, its filter weights are data-driven, which makes RRCNN flexible in the characterization of the local average.
\item[(\romannumeral3)]  In addition to more flexibly computing the local average of the signal, the proposed RRCNN method has all the advantages of deep learning, such as: the non-linear expression ability, the customized loss function according to specific decomposition tasks, etc., which are missing in the traditional signal decomposition methods.
\end{itemize}

Although the proposed RRCNN method is developed based on the IF, its ambition is not to replace the IF, or even to replace any existing signal decomposition method. The RRCNN is essentially a supervised deep learning model. This not only provides all the advantages that the existing signal decomposition methods are lacking, but also gives all the limitations of the supervision model itself. Such as, for any given signal, regardless of the decomposition performance, the existing signal decomposition methods can basically decompose it; however, the RRCNN model does not work without any training based on a set of input signals for which is known the ground truth. In addition, the training process of the RRCNN model can be regarded as the process of exploring the potential patterns of the input signals. If the patterns of the input signals are unique, their decomposition performance will also be greatly reduced.

\subsection{Training process and convergence analysis of RRCNN}
In the training process, the gradient-based back-propagation method is used to learn the filter weights that appear in the convolutional layers by using the training data. For the hyper-parameters in the RRCNN, including the filter lengths and the numbers of neurons in the convolutional layers, Hyperopt\footnote{Github website: https://github.com/hyperopt/hyperopt} is adopted to select their values by using the validation data, which is a Python library for serial and parallel optimization over awkward search spaces for hyper-parameters.

Since when the parameter $\eta$ in Eqn. (\ref{equ::loss_function_TV}) is equal to $0$, it degenerates to the case of Eqn. (\ref{equ::loss_function}), we only discuss the training processes and convergences of the models when their loss functions are given in Eqns. (\ref{equ::loss_function_TV}) and (\ref{equ::loss_function_customized}), respectively.

We first discuss the situation of the RRCNN model equipped with the loss function (\ref{equ::loss_function_TV}). For convenience, we consider that the output of the model only produces one IMF, and the number of recursion is also limited to $1$, that is, $M=1$ and $S_1=1$ in Fig. \ref{fig:CNN}. Suppose that $({X}^{j},{Y}^{j})_{j=1}^J\subset\mathbb{R}^{N}\times\mathbb{R}^{N}$ is a given set of training samples, where $J$ denotes the number of samples. The processe and the loss function of RRCNN are as follows:
\begin{equation}
\label{equ::RRCNN_training}
\begin{split}&\begin{cases}
&{X}_{C1}^{j}=\sigma({X}^{j}\ast {W}_1),\\
&{X}_{C2}^{j}={X}_{C1}^{j}\ast \tilde{W}_2,\\
&\hat{Y}^{j}={X}^{j} - {X}_{C2}^{j},\\
\end{cases}
\end{split}\end{equation}
$L(\hat{Y}, {Y})=\sum_{j=1}^{J}\|\hat{Y}^j-{Y}^j\|_{2}^2+\eta QTV(\hat{Y}^j)$,
where $\tilde{W}_2=softmax(W_2)$, $\sigma=tanh$ in our model, $\eta$ is a non-negative parameter, $W_1\in\mathbb{R}^{K_1}$ and $W_2\in\mathbb{R}^{K_2}$ are the undetermined convolution filters.

According to the gradient descent method and the chain derivation rule, $W_1$ and $W_2$ are learned following the back propagation method, that is,
$W_{h}^{(n+1)}=W_{h}^{(n)}-\lambda\nabla_{W_{h}}L, (h=1,2)$
where $\lambda$ denotes the learning rate, $\nabla_{W_h}L=(\frac{\partial}{\partial {W_{h}}_1},\ldots,\frac{\partial}{\partial {W_{h}}_{K_1}})^{\top}L$, $\frac{\partial L}{\partial {W_{2}}_i}=-\sum_{j=1}^{J}\sum_{t,l=1}^{N}\frac{\partial L}{\partial \hat{Y}^{j}_t}\frac{\partial {{X}^{j}_{C2}}_t}{\partial {\tilde{W_2}_l}}\frac{\partial {\tilde{W_2}_l}}{\partial {W_2}_i}$, $\frac{\partial L}{\partial {W_{1}}_i}=-\sum_{j=1}^{J}\sum_{t,l=1}^{N}\frac{\partial L}{\partial \hat{Y}^{j}_t}\frac{\partial {{X}^{j}_{C2}}_t}{\partial {{X}^{j}_{C1}}_l}\sigma^{\prime}(R^{j}_{l})\frac{\partial R^{j}_{l}}{\partial {{W_1}_i}}$, $\frac{\partial L}{\partial \hat{Y}^{j}_t}=2(\hat{Y}^{j}_t-Y^{j}_t)$, $\frac{\partial {{X}^{j}_{C2}}_t}{\partial {\tilde{W_2}_l}}={{X}^{j}_{C1}}_{(t-\lfloor{K_1}/{2}\rfloor+l)}$, $R^{j}_{l}=({X}^{j}\ast {W}_1)_{l}$, $\frac{\partial R^{j}_{l}}{\partial {{W_1}_i}}={{X}^{j}}_{l-\lfloor{{K_2}}/{2}\rfloor+i}$,
\begin{equation}\nonumber%\begin{split}
\frac{\partial {\tilde{W_2}_l}}{\partial {W_2}_i}=
\begin{cases}
&\frac{exp({{W_2}_i+{W_2}_{l}})}{(\sum_{k}exp({{W_2}_k}))^2}, \text{~if~} l\neq i;\\
& \frac{{exp({{W_2}_i}})\sum_{k\neq i}exp({{W_2}_{k}})}{(\sum_{k}exp({{W_2}_k}))^2}, \text{~otherwise},
\end{cases}%\end{split}
\text{~and~}
\frac{\partial {{X}^{j}_{C2}}_t}{\partial {{X}^{j}_{C1}}_l}=
\begin{cases}
&{\tilde{W_2}_{(l-t+\lfloor{K_1}/{2}\rfloor)}}, \text{~if~} 1\leq l-t+\lfloor{K_1}/{2}\rfloor\leq K_1;\\
&0, \text{~otherwise}.
\end{cases}\end{equation}

Next, we discuss the convergence of the training process of the RRCNN model expressed in Eqn. (\ref{equ::RRCNN_training}).

\begin{thm}\label{thm::RRCNN}
For the RRCNN model defined in Eqn. (\ref{equ::RRCNN_training}), the sequences $\{W_1^{(n)}\}$ and $\{W_2^{(n)}\}$ are generated by the gradient descent algorithm. Then, there exists a positive constant $L$ independent of the input data, so that when the learning rate $\lambda$ is less than $L$, $\{W_1^{(n)}\}$ and $\{W_2^{(n)}\}$ converge to their corresponding stationary points, i.e., the sets $\{W_1: \nabla_{W_1}L={\bf 0}\}$ and $\{W_2: \nabla_{W_2}L={\bf 0}\}$, respectively.
\end{thm}
\begin{pf}
From the Lagrange's mean value theorem, it is obvious to find that the composition function composed by two functions, the $\beta_1$- and $\beta_2$-smooth respectively, is still $\beta$-smooth ($\beta\leq \beta_1\beta_2$) under the condition that it can be composited.

According to Eqn. (\ref{equ::RRCNN}), the RRCNN model can be seen as a composition function composed of a series of functions. Then, combining with Theorem \ref{thm::convergence}, we have that if the function of the $i$-layer in the RRCNN model is $\beta_i$-smooth, it can be proved that the weights sequences obtained by the gradient descent method, converge under the condition that the learning rate satisfies $\lambda\leq {2}/{\Pi \beta_i}$.

For the function $\tilde{W}_2=softmax(W_2)$ that is included in the RRCNN model, the second partial derivative of each ${\tilde{W_{2}}}_{l}$ ($l=1,\ldots K_2$) with respect to its independent variable vector $W_2$ exists and satisfies:
\begin{equation}\nonumber\begin{split}
&\left\vert\frac{\partial^2 {\tilde{W_2}_l}}{\partial {W_2}_i\partial {W_2}_j}\right\vert=\frac{exp({{W_2}_i})}{(\sum_{k}exp({{W_2}_k}))^3}\begin{cases}
 \sum_{k\neq i}exp({{W_2}_{k}})\vert\sum_{k}exp({{W_2}_{k}})-2exp({{W_2}_i})\vert,& \text{if~} l=i=j,\\
exp({{W_2}_l})\vert 2exp({{W_2}_i})-\sum_{k}exp({{W_2}_{k}})\vert,& \text{if~} l\neq i=j,\\
exp({{W_2}_i+{W_2}_{l}+{W_2}_j}),& \text{if~} l\neq i\neq j,\\
\end{cases}\leq 2.
\end{split}\end{equation}
Hence, it states that each ${\tilde{W_{2}}}_{l}$ is $\beta$-smooth (and $\beta\leq 2$) according to the Lagrange's mean value theorem.

For the rest functions involved in RRCNN include quadratic function, $\tanh$, and 1-D convolution operation, these functions can be easily proved to be $\beta$-smooth ($\beta$ here is a general representation, and the $\beta$ value of each function may not be the same.) by judging that their second derivative functions exist and bounded. Therefore, the conclusion is proved.
\end{pf}

For the case of the model with an orthogonal constraint in the loss function $L$ in Eqn. (\ref{equ::loss_function_customized}), the orthogonal constraint is a new obstacle compared to the previous model. However, in the field of optimization, the study of optimization problems with orthogonal constraints has become very common. And the gradient-based projection method can be used to find ${\bf W}^{o}$ with convergence guarantees \cite{ablin2018faster,gao2018new}. Furthermore, under the idea of back propagation used in updating the weights of the neural networks, the solutions of problem (\ref{equ::loss_function_customized}) can be obtained according to Alg. \ref{alg::orthogonal_RRCNN}.

\begin{algorithm}
\caption{RRCNN with orthogonal constraint} \label{alg::orthogonal_RRCNN}
%\KwData{$X\in\mathbb{R}^{N}$, and the number of IMFs $M$}
%\KwResult{the IMFs and residue of $X$}
 $i=0$, given the learning rates $\lambda_1, \lambda_2$, and initialize the ${\bf W}^{o}, {\bf W}$\;
 \While{$i<Max\_Iter$}{
 	${\bf W}^{o}\leftarrow{\bf W}^{o}-\lambda_1\nabla_{{\bf W}^{o}}L$\;
	${\bf W}^{o}\leftarrow\mathcal{P}_{\mathcal{S}_{p,q}}({\bf W}^{o})$, where $\mathcal{P}_{\mathcal{S}_{p,q}}({\bf W}^{o})$ denotes the projection of ${\bf W}^{o}\in\mathbb{R}^{p\times q}$ to the Stiefel manifold $\mathcal{S}_{p,q}$, $\mathcal{P}_{\mathcal{S}_{p,q}}({\bf W}^{o})=\tilde{\bf U}\tilde{\bf V}^{\top}$, and $\tilde{\bf U}\Sigma\tilde{\bf V}^{\top}$ is the reduced singular value decomposition of ${\bf W}^{o}$\;
	${\bf W}\leftarrow{\bf W}-\lambda_2\nabla_{{\bf W}}L$ according to the back propagation method\;
	$i\leftarrow i+1$\;
 }
\end{algorithm}

Since the convergence analysis of $\bf W$ calculated based on the gradient descent method is consistent with Theorem \ref{thm::RRCNN}, and that of ${\bf W}^o$ calculated based on the gradient projection method has been discussed in the literature \cite{ablin2018faster,gao2018new}, so the convergence of Alg. \ref{alg::orthogonal_RRCNN} can be obtained intuitively.

\begin{rem}
Similar to the case with loss function in Eqn. (\ref{equ::loss_function_TV}), we assume that in the RRCNN model, there are only two RRCNN inner loop blocks, i,e., $M=2$ in Fig. \ref{fig:CNN}. Furthermore, the IMFs obtained by the two blocks satisfy orthogonality, i.e., $\Omega_2=\{(1,2)\}$, $\hat{\Omega}_2=\{1,2\}$ and $\hat{\Omega}_2^{c}=\emptyset$ in Eqn. (\ref{equ::loss_function_customized}). In this case, the RRCNN model can be reduced to a simpler formula, which looks close to the expressions of some orthogonal constraint algorithms, which reads:
\begin{equation}%\begin{split}
\label{equ::orthogonal_RRCNN}
\min_{{\bf W},{\bf W}^{o}} \|{\bf W}^{o}\hat{\bf Y}-{\bf Y}\|_{F}^2, \text{s.t. } {\bf W}^{o}{{\bf W}^{o}}^{\top}={\bf I},
%\end{split}
\end{equation}
where $\hat{\bf Y}$ depends on ${\bf W}$.
\end{rem}

\section{Experiments}
\label{sec:experiments}

To evaluate the performance of the proposed RRCNN inner loop block and RRCNN models, we test them against seven aspects, which are: 1) Can RRCNN inner loop block be used to find the local average of the non-stationary signal? 2) Is RRCNN inner loop block still effective on noisy signal? 3) Can RRCNN be used to decompose the non-stationary signals? 4) Can RRCNN be effective on noisy signals? 5) Can RRCNN be effective on the signals composed of orthogonal mono-components? 6) Can RRCNN be effective on solutions of differential equations? 7) Is RRCNN capable of processing real signals? Furthermore, we discuss the computational time of RRCNN, and compare it with other methods. In addition, limited by the length of the paper, we submit some experiments as supplementary materials.

In the experiments, we divide each of the constructed input data into the training and validation datasets with ratio $7:3$. Moreover, our proposed signal average method, i.e., the RRCNN inner loop block, is compared with the existing signal average methods based on cubic spline envelopes (CSE) \cite{huang1998empirical}, optimization model (OP), a segment power-function based envelopes (SPFE) \cite{huang1999new} and iterative filtering (IF) \cite{cicone2016adaptive}, respectively. For simplicity, we denote the averages obtained from the CSE, OP, SPFE, and IF as CSA, OPA, SPFA, and IF respectively. And the RRCNN\footnote{Code of RRCNN is available at https://github.com/zhoudafa08/RRCNN} method will be compared with the state-of-the-art methods, including EMD, IF\footnote{Code of IF: http://people.disim.univaq.it/$\sim$antonio.cicone/Software.html}, VMD\footnote{Code of VMD: https://www.mathworks.com/help/wavelet/ref/vmd.html}\cite{dragomiretskiy2013variational}, continuous wavelet transform based synchrosqueezing (SYNSQ\_CWT) \cite{daubechies2011synchrosqueezed}, short time Fourier transform based synchrosqueezing (SYNSQ\_STFT\footnote{Codes of SYNSQ\_CWT and SYNSQ\_STFT: https://github.com/ebrevd o/synchrosqueezing}) \cite{daubechies2011synchrosqueezed}, INCMD\footnote{Code of INCMD: https://github.com/sheadan/IterativeNCMD} \cite{tu2020iterative}, EWT\footnote{Code of EWT: https://ww2.mathworks.cn/help/wavelet/ug/empirical-wave let-transform.html} \cite{gilles2013empirical}, FDM\footnote{Code of FDM: https://www.researchgate.net/publication/2745702 45\_Matl ab\_Code\_Of\_The\_Fourier\_Decomposition\_Method\_FDM} \cite{singh2017fourier}, and its variant called DCT\_GAS\_FDM\footnote{Code of DCT\_GAS\_FDM: https://www.researchgate.net/publication/32629 4577\_MATLABCodeOfFDM\_DCT\_DFT\_FIR\_FSASJuly2018} \cite{singh2018novel}. In addition, in the experiments for verifying the robustness of noise interference, the proposed RRCNN method is compared with the ensemble EMD (called EEMD\footnote{Code of EEMD: http://perso.ens-lyon.fr/patrick.flandrin/emd.html}) model {\cite{wu2009ensemble}}; and it is compared with the M-LFBF\footnote{Code of M-LFBF: http://perso.ens-lyon.fr/nelly.pustelnik/}\cite{pustelnik2012multicomponent, pustelnik2014empirical} model to verify the orthogonality of the decomposed components.

\begin{table}[H]\scriptsize
\setlength\tabcolsep{2pt}
%\begin{adjustwidth}{0in}{0in}
\centering
\caption{Evaluation indices}
\renewcommand\arraystretch{1.2}
\begin{tabular}{cccc}
\toprule
Metric &  MAE & RMSE & $\rho(c_1,c_2)$ \\
\midrule
Expression & $\frac{1}{N}\sum_{t=1}^{N}|\hat{Y}_t-Y_t|$ & $\sqrt{\frac{1}{N}\sum_{t=1}^{N}(\hat{Y}_t-Y_t)^2}$ & $\frac{|\langle c_1, c_2\rangle|}{\|C_1\|_2\|c_2\|_2}$  \\
\bottomrule
\end{tabular}
\centering
\label{tab::metric}
%\end{adjustwidth}
\end{table}

The results are measured by the metrics listed in Tab. \ref{tab::metric}, where Mean Absolute Error (MAE) and Root Mean Squared Error (RMSE) are used to measure the errors between the predicted results $\hat{Y}\in\mathbb{R}^{N}$ and the ground truth $Y\in\mathbb{R}^{N}$, and $\rho(c_1, c_2)$ is used to evaluate the orthogonality between the resulting components $c_1\in\mathbb{R}^{N}$ and $c_2\in\mathbb{R}^{N}$.

\subsection{Can RRCNN inner loop block be used to find the local average of the non-stationary signal?}
\label{subsec:1}
We first evaluate the performance of the proposed RRCNN inner loop block in solving the local average of the non-stationary signal. Several signals composed of the linear function and the mono-component function, or just the mono-component function,  are constructed as the inputs, where the mono-component function can generally be expressed as $a(t)\cos \theta(t)$, which meets $a(t),\theta^{\prime}(t)>0$ $\forall t$, and the changes in time of $a(t)$ and $\theta^{\prime}(t)$ are much slower than that of $\theta(t)$. Ideally, the average of the mono-component signal is zero. The input signals and the corresponding labels we construct in this part are listed in Tab. \ref{tab::inputs1}. It should be pointed out that the label here represents the first IMF of the corresponding input signal, not the local average. The local average can be computed by subtracting the label from the input signal.

After the RRCNN inner loop block is trained with the inputs in Tab. \ref{tab::inputs1}, we select three mono-component signals with different instantaneous frequencies and instantaneous amplitudes, discussed in Examples 1-2, which are not included in the inputs, to test the performance of the RRCNN inner loop block, respectively.

\begin{table}[H]\scriptsize
%\setlength\tabcolsep{0.1pt}
%\begin{adjustwidth}{0in}{0in}
\centering
\caption{Inputs and labels used in Section \ref{subsec:1}, where $t\in[0,3]$.}
\renewcommand\arraystretch{1.2}
\begin{tabular}{ccccc}
\toprule
$x_1(t)$&  $x_2(t)$ & Inputs & Labels & Notes\\
\midrule
\multirow{2}{*}{$0.1kt$} & $\cos(3lt)$ & \multirow{4}{*}{$x_1(t)+x_2(t)$} & \multirow{4}{*}{$x_2(t)$}  & \multirow{2}{*}{$k=2,3,\ldots, 9$} \\
& $\cos(3klt+t+\cos(t))$ &  & &   \\
\multirow{2}{*}{$0$} & $\cos(3lt)$ & & & \multirow{2}{*}{$l=2,4,6,8$}\\
& $\cos(3klt+t+\cos(t))$ & &  &  \\
\midrule
\multirow{2}{*}{$0.1k$} & $\sin(3klt)$ &  \multirow{4}{*}{$x_1(t)+x_2(t)$} & \multirow{4}{*}{$x_2(t)$}  &  \multirow{2}{*}{$k=1,2,\ldots,10$}\\
& $\sin(3klt+t^2+\cos(t))$  &  &  &   \\
\multirow{2}{*}{$0$} & $\sin(3klt)$ & & & \multirow{2}{*}{$l=2,4,\ldots, 28$}   \\
& $\sin(3klt+t^2+\cos(t))$ &  &  &    \\
\midrule
\multirow{2}{*}{$3+2\cos(0.5kt)$} & $\cos(0.5klt^2)$ &  \multirow{4}{*}{$x_1(t)x_2(t)$} & \multirow{4}{*}{$x_1(t)x_2(t)$} & \multirow{2}{*}{$k=2,3,\ldots,6$} \\
& $\cos(0.5lt^2+l\cos(t))$  &  &  &   \\
\multirow{2}{*}{1.0} & $\cos(0.5klt^2)$ & & & \multirow{2}{*}{$l=4,5,\ldots, 9$}  \\
& $\cos(0.5lt^2+l\cos(t))$  &  &  &  \\
\bottomrule
\end{tabular}
\centering
\label{tab::inputs1}
%\end{adjustwidth}
\end{table}

{\bf Example 1}: $x(t)=(3+2\cos(2t))\cos(2t^2)$, $t\in[0,3]$.

\begin{table}[H]\scriptsize
\begin{adjustwidth}{0in}{0in}
\centering
\caption{Metrics of the moving averages of the input in Example 1.}
\renewcommand\arraystretch{1.2}
\begin{tabular}{cccccc}
\toprule
Metric &  CSA & OPA & SPFA & IF & {\bf RRCNN} \\
\midrule
MAE & 0.4082 & 0.5296 & 0.6558 & 1.0977 & {\bf 0.1656}  \\
RMSE & 0.6001 & 0.8763 & 0.8884 & 1.3829 & {\bf 0.2040} \\
\bottomrule
\end{tabular}
\centering
\label{tab::avg_eg1}
\end{adjustwidth}
\end{table}

\begin{figure}[!t]
%\centering
\subfigure[Example 1]{
\begin{minipage}[t]{0.33\linewidth}%\centering
\includegraphics[width=2.5in]{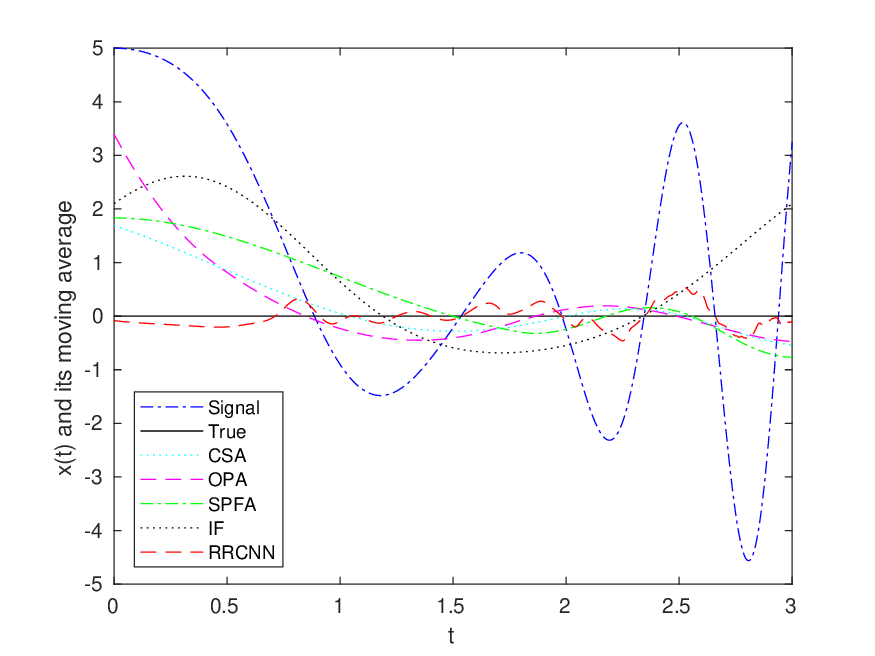}
\end{minipage}
}
\subfigure[Example 2]{
\begin{minipage}[t]{0.33\linewidth}%\centering
\includegraphics[width=2.5in]{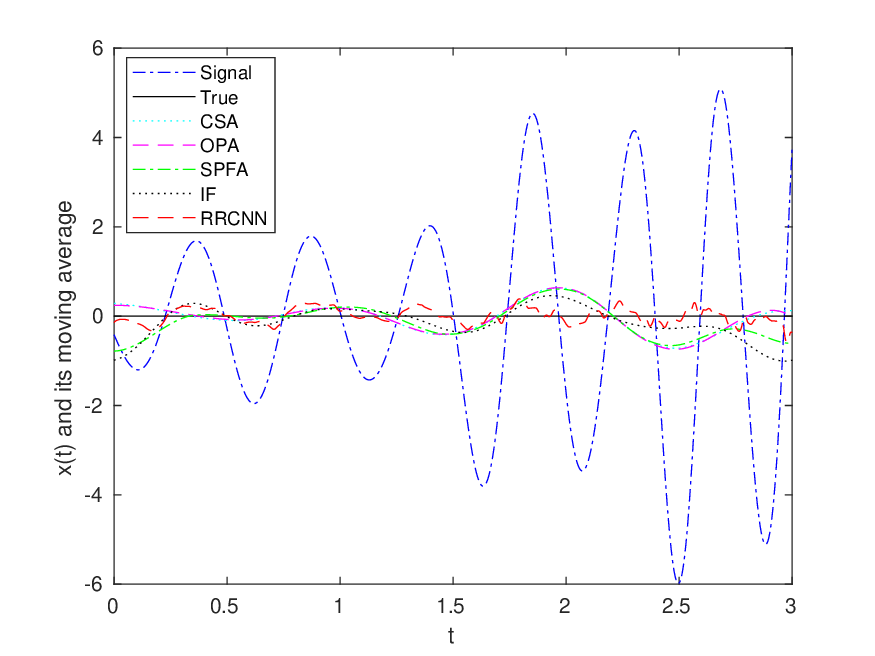}
\end{minipage}
}
\subfigure[Example 3]{
\begin{minipage}[t]{0.33\linewidth}%\centering
\includegraphics[width=2.5in]{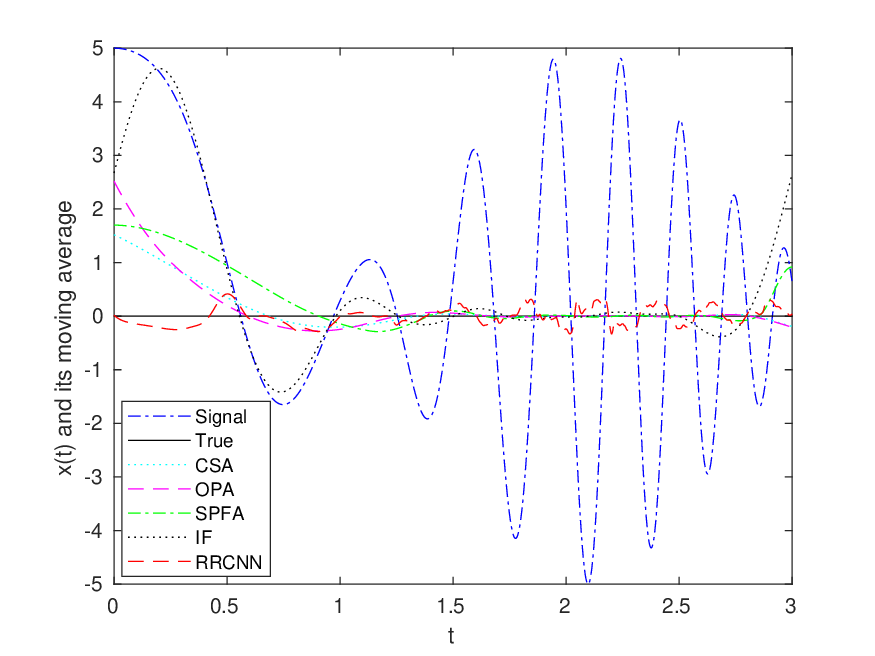}
\end{minipage}
}
\caption{Moving averages by different methods of Examples 1-3.}
\label{fig:avg_all}
\end{figure}

{\bf Example 2}: $x(t)=(2t+\cos(2t^2))\cos(12t+t^2+2\cos(t))$, $t\in[0,3]$.

\begin{table}[H]\scriptsize
\begin{adjustwidth}{0in}{0in}
\centering
\caption{Metrics of the moving averages of the input in Example 2.}
\renewcommand\arraystretch{1.2}
\begin{tabular}{cccccc}
\toprule
Metric &  CSA & OPA & SPFA & IF & {\bf RRCNN} \\
\midrule
MAE& 0.2544 & 0.2514 & 0.2943 & 0.2791 & {\bf 0.1442}  \\
RMSE& 0.3267 & 0.3276  & 0.3662 & 0.3672 & {\bf 0.1718}\\
\bottomrule
\end{tabular}
\centering
\label{tab::avg_eg2}
\end{adjustwidth}
\end{table}

{\bf Example 3}: $x(t)=(3+2\cos(3t))\cos(5t^2)$, $t\in[0,3]$.

\begin{table}[H]\scriptsize
\begin{adjustwidth}{0in}{0in}
\centering
\caption{Metrics of the moving averages of the input in Example 3.}
\renewcommand\arraystretch{1.2}
\begin{tabular}{cccccc}
\toprule
Metric &  CSA & OPA & SPFA & IF & {\bf RRCNN} \\
\midrule
MAE& 0.2067 & 0.2448 & 0.3646 & 0.8624 & {\bf 0.1418}  \\
RMSE& 0.4185 & 0.5427 & 0.6388 & 1.5833 & {\bf 0.1712} \\
\bottomrule
\end{tabular}
\centering
\label{tab::avg_eg3}
\end{adjustwidth}
\end{table}

The moving averages of the signals in Examples 1-3 obtained from different methods are shown in Fig. \ref{fig:avg_all} (a)-(c), respectively, and the errors between the obtained moving averages and the true average are listed in Tabs. \ref{tab::avg_eg1}-\ref{tab::avg_eg3}, respectively. According to the results, we can observe the following phenomena:
\begin{itemize}
\item [(\romannumeral1)] The existing methods are prone to boundary effects, which can be seen from the left boundaries of Fig. \ref{fig:avg_all} (a)-(c). However, the RRCNN inner loop block method can avoid this problem to a certain extent.
\item [(\romannumeral2)] When the signal is in a situation where the amplitude changes quickly and the frequency changes slowly, the RRCNN inner loop block performs best among all models according to the left parts of Fig. \ref{fig:avg_all} (a)-(c). When the amplitude change is reduced and the frequency change is accelerated, its performance may even be inferior to other models, which can be seen from the right half of Fig. \ref{fig:avg_all} (c).
\item [(\romannumeral3)] Even though the RRCNN inner loop block has some dazzling and bleak performance compared with other local average methods, it can be seen from Tabs. \ref{tab::avg_eg1}-\ref{tab::avg_eg3} that the MAE and RMSE of RRCNN are significantly reduced compared to other models.
\end{itemize}

The reason for the phenomenon above can be attributed to:
\begin{itemize}
\item [(\romannumeral1)] The averages obtained by the comparison methods are basically determined by the local information of the signal, which makes the results reasonable when the information is sufficient (e.g., the part of the amplitude change is reduced and the frequency change is accelerated); and the results differ greatly when the information is insufficient (e.g., the part of the amplitude changes quickly and the frequency changes slowly).
\item [(\romannumeral2)]  The filter weights of each convolutional layer in the RRCNN are shared, they are determined by all the information contained in the whole signal. Therefore, the average obtained by the RRCNN is relatively stable, and it is not easy to cause interference due to the large difference in the changing of the signal amplitude and frequency.
\end{itemize}

\subsection{Is RRCNN inner loop block still effective on noisy signal?}
\label{subsec:2}
In this part, we consider the robustness of the RRCNN inner loop block model to noise based on the constructed inputs and labels in Section \ref{subsec:1}. Specifically, each input signal is perturbed with additive Gaussian noise with the signal-to-noise ratio (SNR) set to $15dB$, and the corresponding label remains unchanged, as detailed in Tab. \ref{tab::inputs2}. Similar to the section above, we select a noisy signal, which is essentially the signal in Examples 2-3 with additive Gaussian noise, and is described in Examples 4-5, to test the performance.

\begin{table}[H]\scriptsize
\begin{adjustwidth}{0in}{0in}
\centering
\caption{Inputs disturbed by the Gaussian noise with the SNR set to $15dB$, and the labels used in Section \ref{subsec:2}, where $t\in[0,3]$.}
\renewcommand\arraystretch{1.2}
\begin{tabular}{ccccc}
\toprule
$x_1(t)$&  $x_2(t)$ & Inputs & Labels & Notes\\
\midrule
\multirow{2}{*}{$0.1kt$} & $\cos(3lt)$ &\multirow{4}{*}{$x_1(t)+x_2(t)+\varepsilon(t)$} & \multirow{4}{*}{$x_2(t)$}  & \multirow{2}{*}{$k=2,3,\ldots,9$} \\
& $\cos(3klt+t+\cos(t))$ & & &   \\
\multirow{2}{*}{$0$} & $\cos(3lt)$ & & & \multirow{2}{*}{$l=2,4, 6, 8$}\\
& $\cos(3klt+t+\cos(t))$ & &  &  \\
\midrule
\multirow{2}{*}{$0.1k$} & $\sin(3klt)$ &   \multirow{4}{*}{$x_1(t)+x_2(t)+\varepsilon(t)$} & \multirow{4}{*}{$x_2(t)$}  &  \multirow{2}{*}{$k=1,2,\ldots,10$}\\
& $\sin(3klt+t^2+\cos(t))$  &  &  &   \\
\multirow{2}{*}{$0$} & $\sin(3klt)$ & & & \multirow{2}{*}{$l=2,4,\ldots, 28$}   \\
& $\sin(3klt+t^2+\cos(t))$ &  &  &    \\
\midrule
\multirow{2}{*}{$3+2\cos(0.5kt)$} & $\cos(0.5klt^2)$ &  \multirow{4}{*}{$x_1(t)x_2(t)+\varepsilon(t)$}  & \multirow{4}{*}{$x_1(t)x_2(t)$} & \multirow{2}{*}{$k=2,3,\ldots,6$} \\
& $\cos(0.5lt^2+l\cos(t))$  &  &  &   \\
\multirow{2}{*}{1.0} & $\cos(0.5klt^2)$ & & & \multirow{2}{*}{$l=4,5,\ldots, 9$}  \\
& $\cos(0.5lt^2+l\cos(t))$  &  &  &   \\
\bottomrule
\end{tabular}
\centering
\label{tab::inputs2}
\end{adjustwidth}
\end{table}

{\bf Example 4}: $x(t)=(2t+\cos(2t^2))\cos(20t+t^2+2\cos(t))+\varepsilon(t)$, where $t\in[0,3]$ and $SNR=15dB$.

\begin{table}[H]\scriptsize
\begin{adjustwidth}{0in}{0in}
\centering
\caption{Metrics of the moving averages of the input in Example 4 obtained from different methods.}
\renewcommand\arraystretch{1.2}
\begin{tabular}{cccccc}
\toprule
Metric &  CSA & OPA & SPFA & IF & {\bf RRCNN} \\
\midrule
MAE & 0.2023 & 0.1998 &0.1936 &0.2183 & \textbf{0.1190}  \\
RMSE &  0.2529 & 0.2501 & 0.2420 & 0.2762 & \textbf{0.1500} \\
\bottomrule
\end{tabular}
\centering
\label{tab::avg_eg4}
\end{adjustwidth}
\end{table}

{\bf Example 5}: $x(t)=(3+2\cos(3t))\cos(5t^2)+\varepsilon(t)$, where $t\in[0,3]$ and $SNR=15dB$.

\begin{figure}[H]
\centering
\subfigure[Example 4]{
\begin{minipage}[t]{0.48\linewidth}%\centering
\includegraphics[scale=0.45]{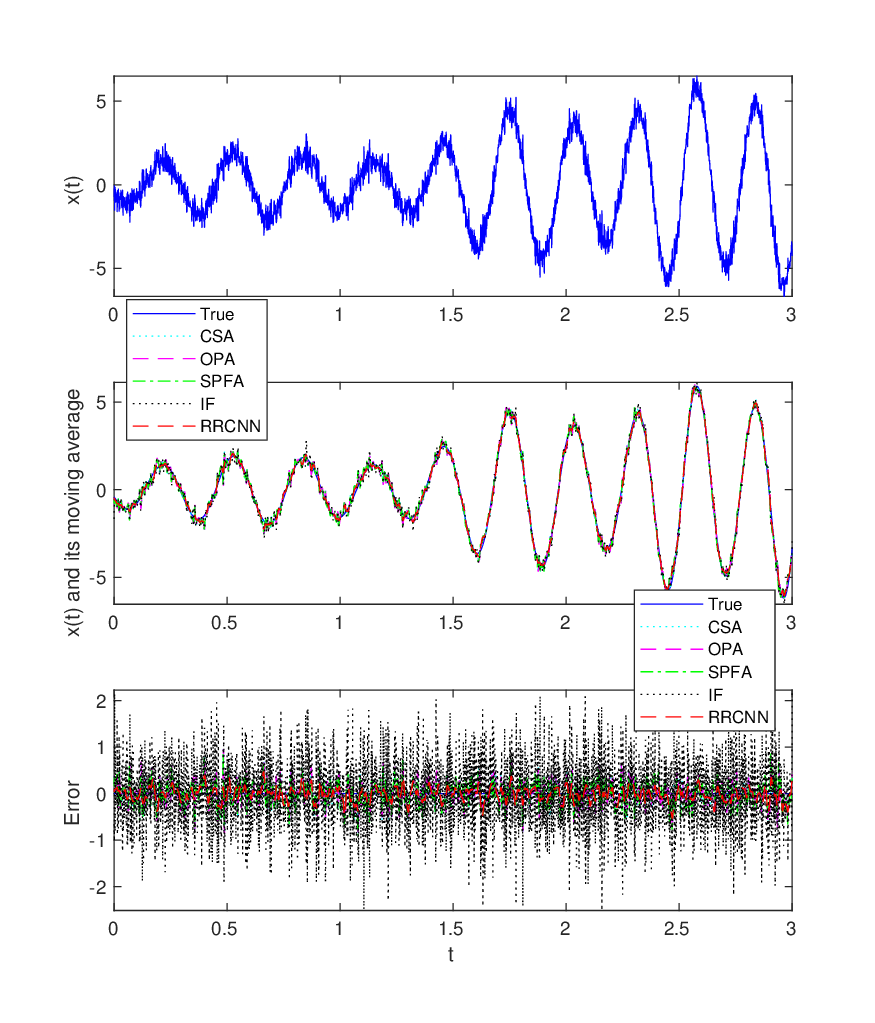}
\end{minipage}
}
\subfigure[Example 5]{
\begin{minipage}[t]{0.48\linewidth}%\centering
\includegraphics[scale=0.47]{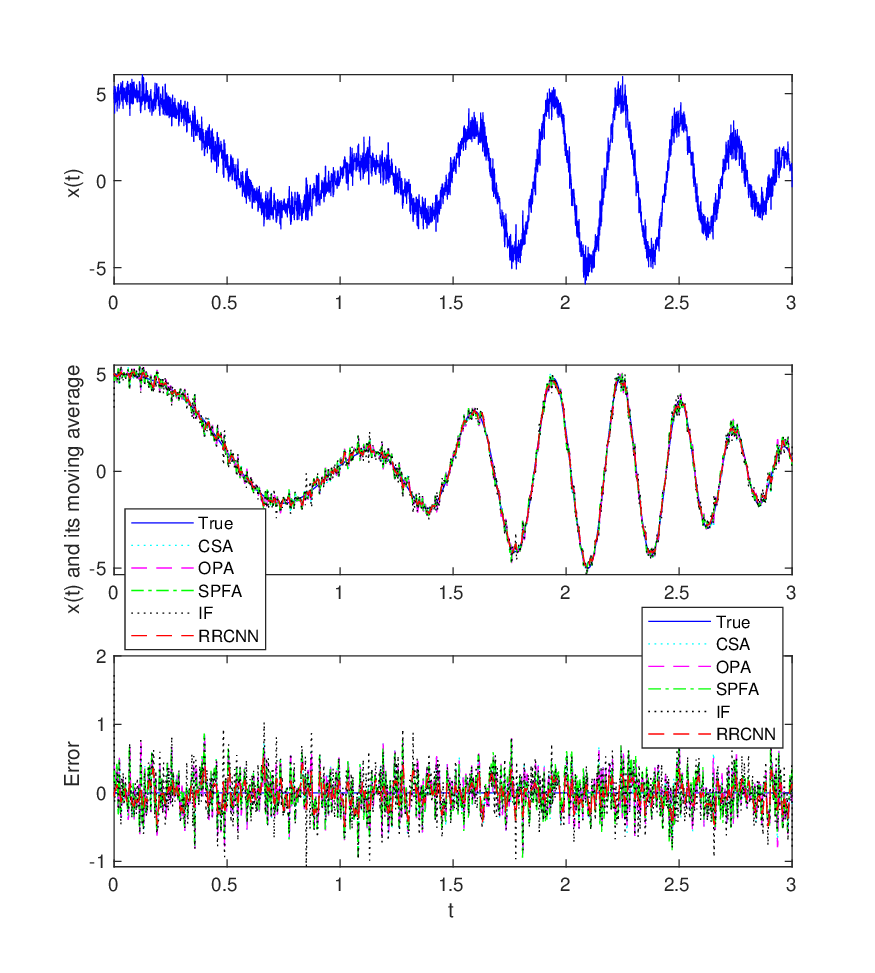}
\end{minipage}
}
\caption{Moving averages by different methods of Examples 4-5.}
\label{fig:avg_eg4-5}
\end{figure}

The results of Examples 4-5 are shown in Fig. \ref{fig:avg_eg4-5} (a)-(b) and Tabs. \ref{tab::avg_eg4}-\ref{tab::avg_eg5}. From the results, we can find that the RRCNN inner loop block performs the the most robust among all models for the signal with additive Gaussian noise. Specifically, in Example 4, the RRCNN inner loop block reduces the MAE and RMSE from $0.1936, 0.2420$, for the second the best method (i.e., SPFA), to $0.1190, 0.1500$, respectively; and in Example 5, the RRCNN inner loop block reduces the MAE and RMSE from $0.2004, 0.2520$, for the second the best method (i.e., SPFA), to $0.1290, 0.1623$, respectively.

\begin{table}[H]\scriptsize
\begin{adjustwidth}{0in}{0in}
\centering
\caption{Metrics of the moving averages of the input in Example 5.}
\renewcommand\arraystretch{1.2}
\begin{tabular}{cccccc}
\toprule
Metric &  CSA & OPA & SPFA & IF & {\bf RRCNN} \\
\midrule
MAE& 0.2141 & 0.2126  &0.2004 &0.2339 & {\bf 0.1290} \\
RMSE& 0.2676 & 0.2658 & 0.2520 & 0.2967& {\bf 0.1623}  \\
\bottomrule
\end{tabular}
\centering
\label{tab::avg_eg5}
\end{adjustwidth}
\end{table}

\subsection{Can RRCNN be used to decompose the non-stationary signals?}
\label{subsec:3}

After evaluating the proposed RRCNN inner loop block model in calculating the local average, the next task is to assess the decomposition performance of RRCNN for non-stationary signals. To demonstrate the problem, we only consider decomposing the signals consisting of two components. The input signals in the part can be divided into two categories: one is composed of a mono-component signal and a zero signal, and the other is of two mono-components with close frequencies. The former is to train RRCNN describe the zero local average of the mono-component signal; and the latter is to enable RRCNN to decompose signals with close frequencies, which is the main factor causing the mode mixing effect. The inputs and labels are constructed and shown in Tab. \ref{tab::inputs3}. To challenge the proposed model, we hereby choose the signal composed of two cosine signals with frequencies of $2.5$Hz and $3.4$Hz respectively, described in Example 4, that is more prone to introduce the mode mixing effect in the existing methods.

\begin{table}[H]\scriptsize
%\setlength\tabcolsep{1pt}
%\begin{adjustwidth}{0in}{0in}
\centering
\caption{Inputs and labels used in Section \ref{subsec:3}, where $t\in[0,6]$.}
\renewcommand\arraystretch{1.2}
\begin{tabular}{ccccc}
\toprule
$x_1(t)$&  $x_2(t)$ & Inputs & Labels & Notes\\
\midrule
\multirow{2}{*}{$\cos(k\pi t)$} & $\cos((k+1.5)\pi t)$ &  \multirow{4}{*}{$x_1(t)+x_2(t)$} &  \multirow{4}{*}{$[x_2(t),x_1(t)]$} & \multirow{4}{*}{$k=5,6,\ldots,14$}  \\
& $\cos((k+1.5)\pi t+t^2+\cos(t))$ &  & &    \\
\multirow{2}{*}{$0$} & $\cos((k+1.5)\pi t)$ & & & \\
& $\cos((k+1.5)\pi t+t^2+\cos(t))$& & &  \\
\midrule
\multirow{2}{*}{$\cos(k\pi t)$} & $\cos(kl\pi t)$ &  \multirow{4}{*}{$x_1(t)+x_2(t)$} & \multirow{4}{*}{$[x_2(t),x_1(t)]$}  & \multirow{2}{*}{$k=5,6,\ldots,14$}  \\
& $\cos(kl\pi t+t^2+\cos(t))$ &  &  &    \\
\multirow{2}{*}{$0$} & $\cos(kl\pi t)$ & & & \multirow{2}{*}{$l=2,3,\ldots, 19$}  \\
& $\cos(kl\pi t+t^2+\cos(t))$ & &&   \\
\bottomrule
\end{tabular}
\centering
\label{tab::inputs3}
%\end{adjustwidth}
\end{table}

{\bf Example 6}: $x(t)=\cos(5\pi t)+\cos(6.8\pi t)$, $t\in[0,6]$.

\begin{figure*}[!t]
\centering
\begin{minipage}[t]{0.48\textwidth}\centering
\subfigure[True and obtained components of the signal (top panel) in Example 6. Left of the 2nd-3rd rows: true (green solid curve) and components by EMD (blue dotted curve), INCMD (black dashed curve), and RRCNN (red dot dash curve). Right of the 2nd-3rd rows: the errors between the ground truth and the obtained components by EMD (blue dotted curve), INCMD (black dashed curve), and RRCNN (red dot dash curve), respectively.]{
\includegraphics[height=2.5in, width=3.3in]{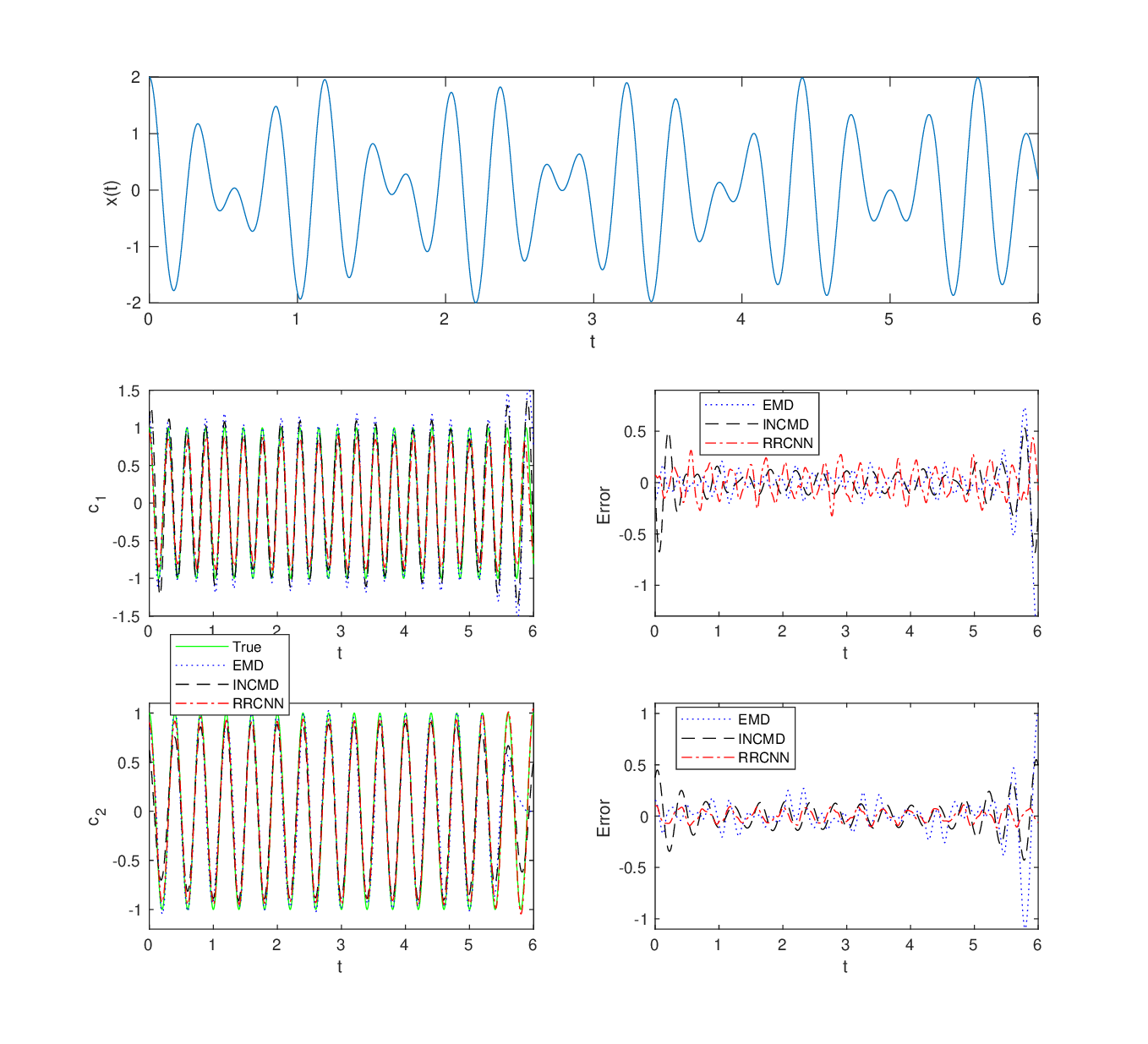}
}
\subfigure[TFE representations of the true and obtained components of the signal in Example 6. Left of the top row: TFE of the ground truths. Right of the top row: TFE of the results by EMD. Left of the bottom row: TFE of the results by INCMD. Right of the bottom row: TFE of the results by RRCNN.]{
\includegraphics[height=1.3in, width=3.3in]{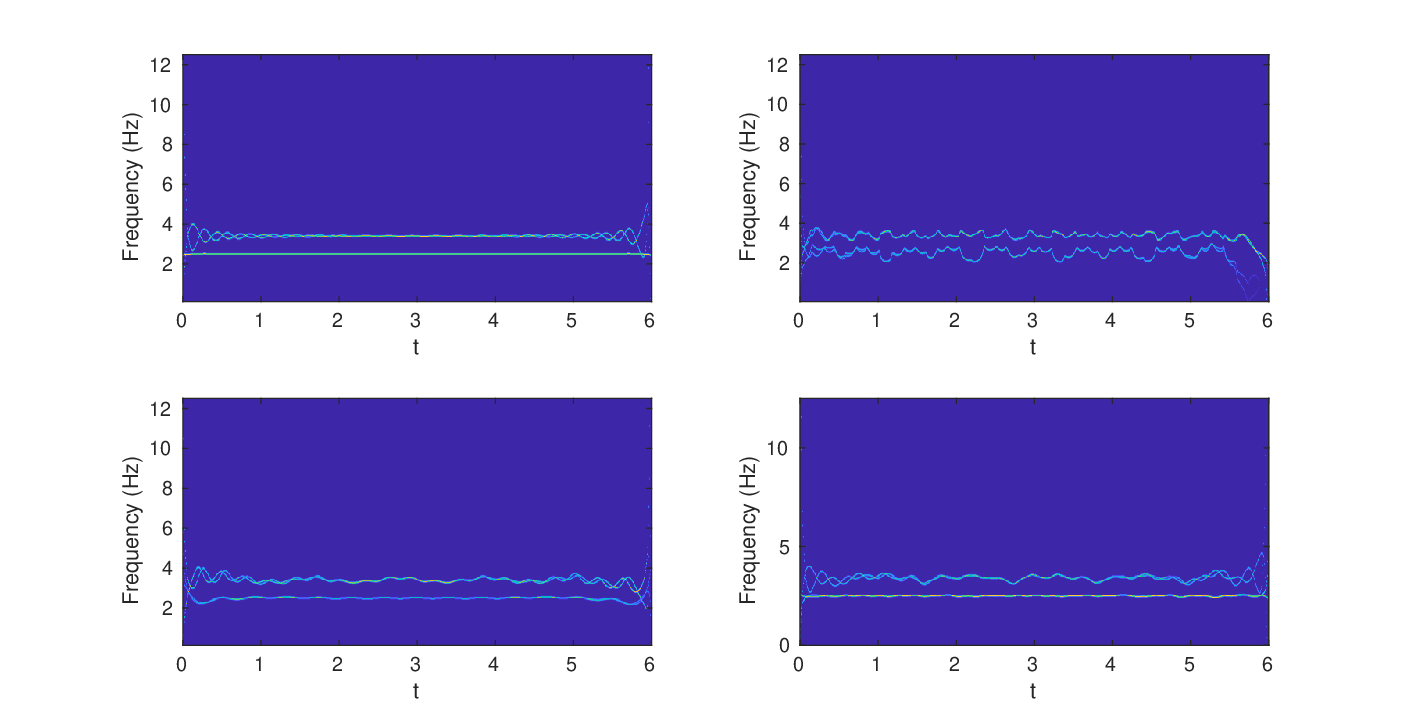}
}
\subfigure[Same as (d) for Example 8.]{
\includegraphics[height=1.9in, width=3.1in]{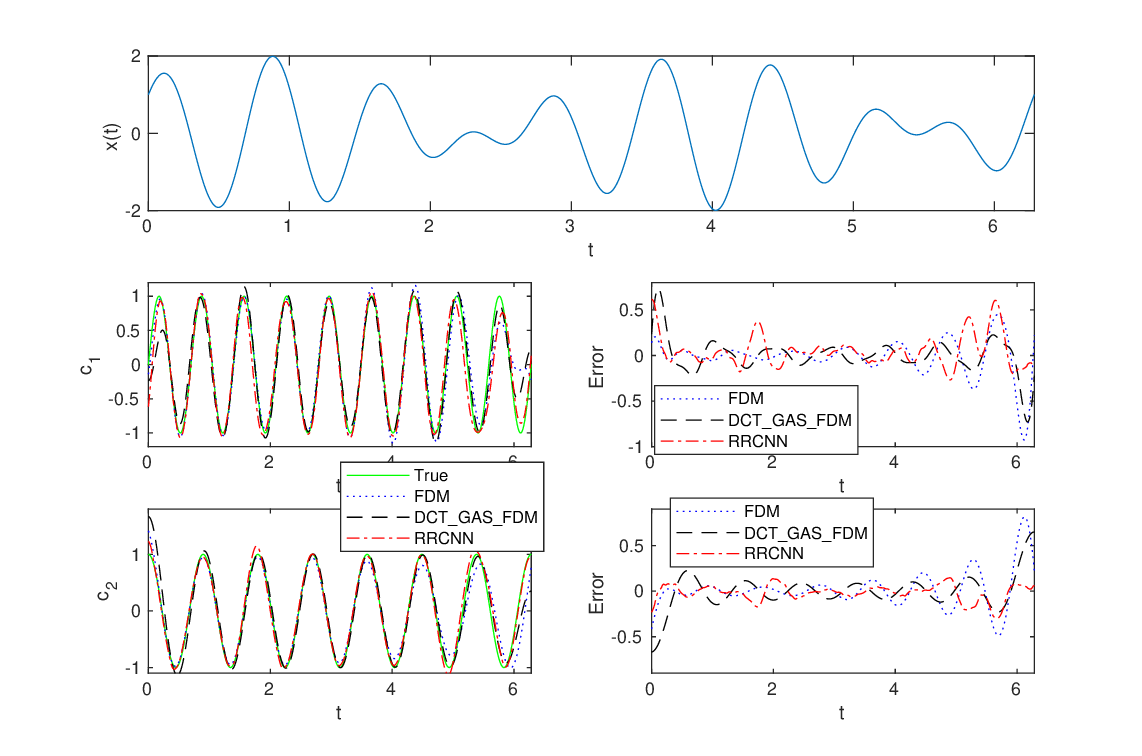}
}
\end{minipage}
\quad
\begin{minipage}[t]{0.48\textwidth}\centering
\subfigure[True and obtained components of the signal (top panel) in Example 7. Left of the 2nd-3rd rows: true (green solid curve) and component by FDM (blue dotted curve), DCT\_GAS\_FDM (black dashed curve), and RRCNN (red dot dash curve). Right of the 2nd-3rd rows: the errors between the  true and obtained components by FDM (blue dotted curve), DCT\_GAS\_FDM (black dashed curve), and RRCNN (red dot dash curve), respectively.]{
\includegraphics[height=2.5in, width=3.3in]{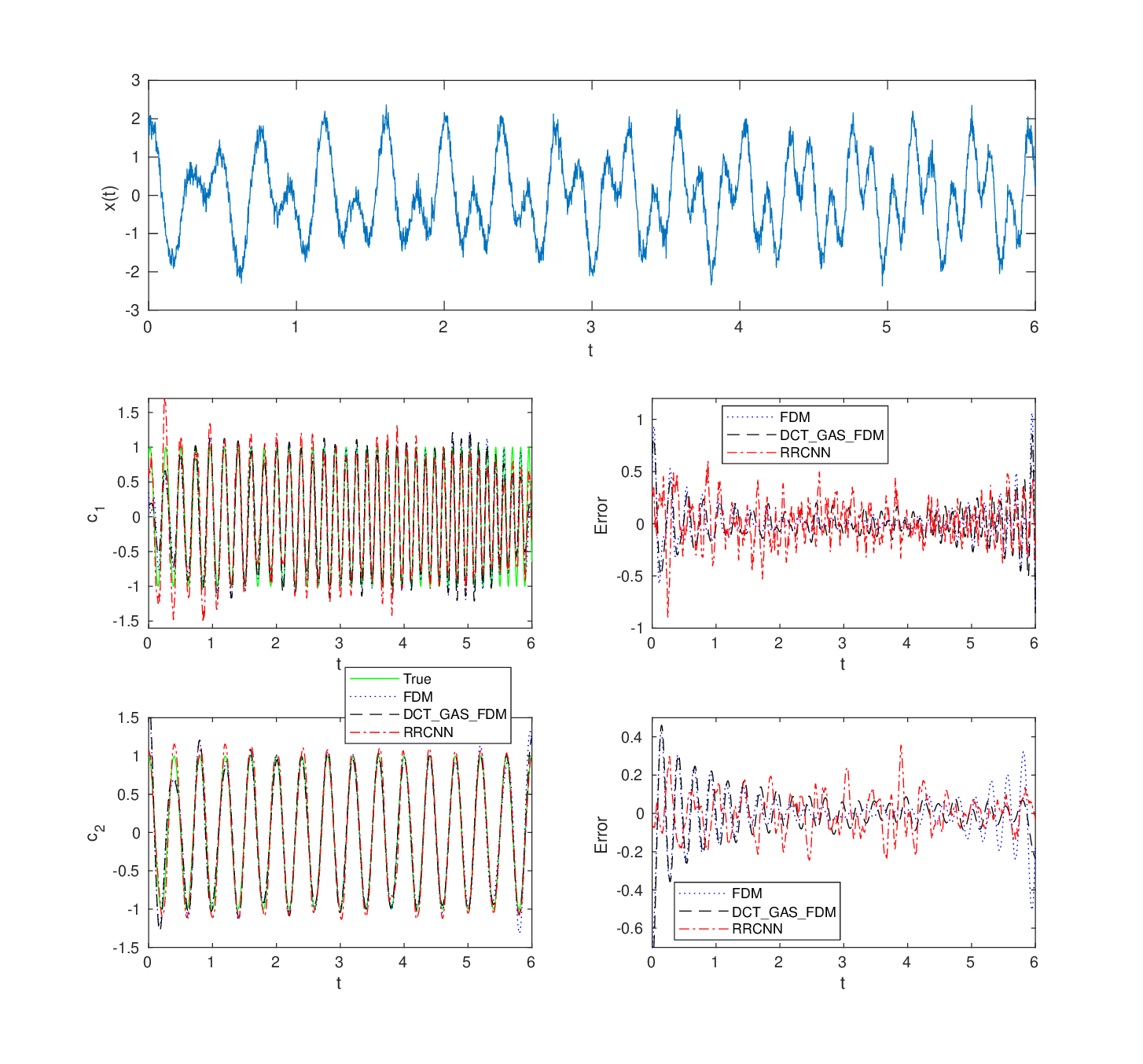}
}
\subfigure[TFE representations of the true and obtained components of the signal in Example 7. Left of the top row: TFE of the ground truths. Right of the top row: TFE of the results by FDM. Left of the bottom row: TFE of the results by DCT\_GAS\_FDM. Right of the bottom row: TFE of the results by RRCNN.]{
\includegraphics[height=1.3in, width=3.3in]{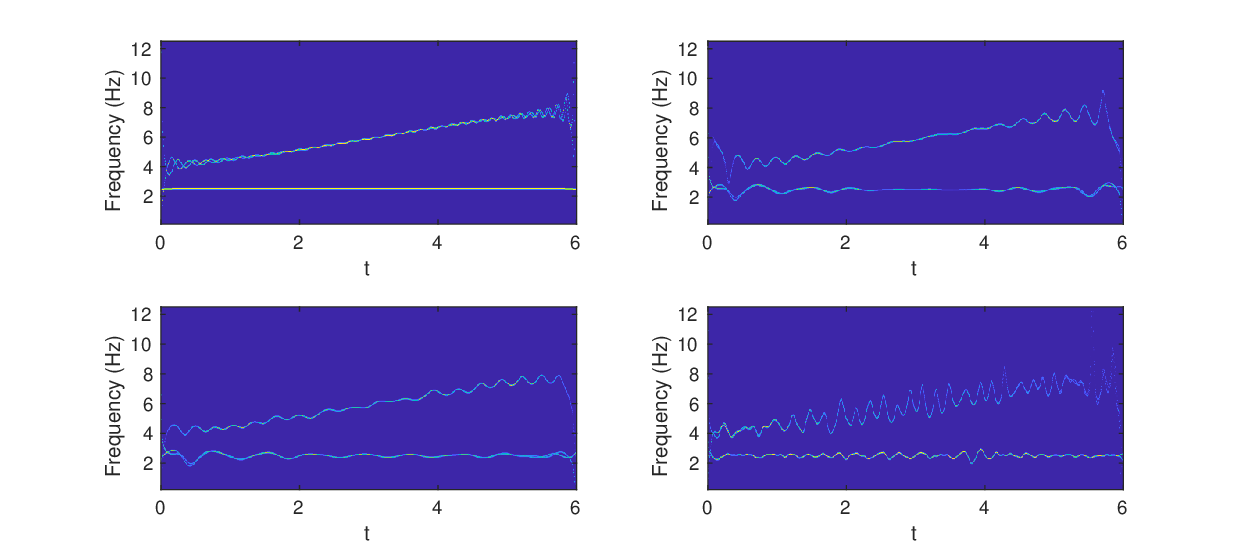}
}
\subfigure[Same as (e) for Example 8.]{
\includegraphics[height=1.35in, width=3.3in]{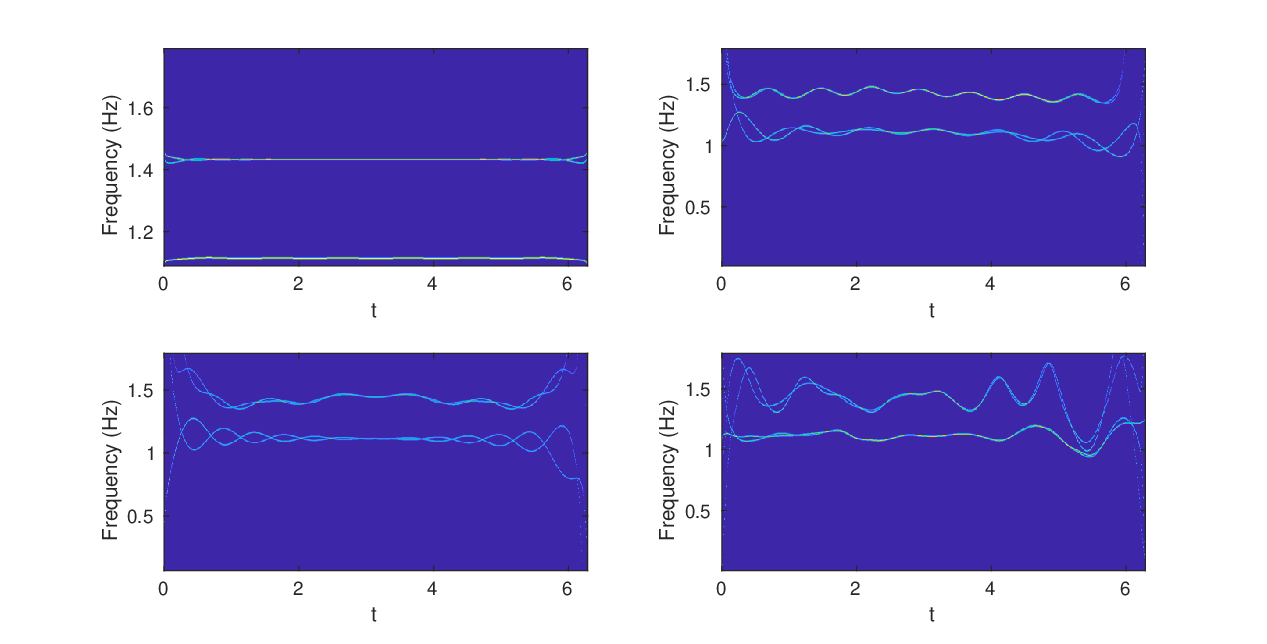}
}
\end{minipage}
\caption{Results of Examples 6-8.}
\label{fig:signal_dec_all}
\end{figure*}

In Example 6, the components predicted by the trained RRCNN model are compared with those obtained from the state-of-the-art methods in signal decomposition. The metrics of the errors between the obtained components and the labels, measured by MAE and RMSE, are shown in Tab. \ref{tab::signal_dec_eg6}. In addition, to compare RRCNN and the existing methods more intuitively, we select the top three methods with comprehensive performance from Tab. \ref{tab::signal_dec_eg6}, i.e., RRCNN, EMD, and INCMD, and plot their obtained components and the corresponding time-frequency-energy (TFE) representations in Fig. \ref{fig:signal_dec_all} (a), (b), respectively. It should be noted that the identification of an optimal TFE representation is a research topic on its own, and it  is out of the scope of this work. Here, we set as TFE representation the Fourier quadrature transforms that was proposed in \cite{singh2018novel}.

\begin{table}[H]\scriptsize
\begin{adjustwidth}{0in}{0in}
\centering
\caption{Metrics of the errors between the obtained components by different methods and the ground truth of Example 6.}
\renewcommand\arraystretch{1.2}
\begin{tabular}{ccccc ccccc}
\toprule
\multirow{2}{*}{Method} &  \multicolumn{2}{c}{$c_1$} & \multicolumn{2}{c}{$c_2$} & \multirow{2}{*}{Method} &  \multicolumn{2}{c}{$c_1$} & \multicolumn{2}{c}{$c_2$} \\
\cline{2-5} \cline{7-10}
& MAE & RMSE &  MAE & RMSE &  & MAE & RMSE &  MAE & RMSE \\
\midrule
EMD\cite{huang1998empirical}& 0.1049 & 0.2110 & 0.1132 & 0.2058 & INCMD\cite{tu2020iterative} &0.1144 & 0.1667 & 0.1160  & 0.1509\\
VMD\cite{dragomiretskiy2013variational}& 0.2193 & 0.2479  &0.2193 &0.2479  & SYNSQ\_CWT\cite{daubechies2011synchrosqueezed} & 0.2227 & 0.2496 & 0.2589 & 0.2965\\
EWT\cite{gilles2013empirical} & 0.2145 &0.2605 &0.2145 &0.2605 & SYNSQ\_STFT\cite{daubechies2011synchrosqueezed} &0.6620 &0.7402 &0.6778 &0.7608 \\
FDM\cite{singh2017fourier} & 0.2438 & 0.2947 & 0.2429 & 0.2893 & DCT\_GAS\_FDM\cite{singh2018novel} & 0.1608 &0.2005 &0.1605 &0.2003 \\
IF\cite{cicone2016adaptive}&0.2253  & 0.2756 &0.2060 & 0.2489  & \textbf{RRCNN}&  \textbf{0.1013} & \textbf{0.1242} & \textbf{0.0331} & \textbf{0.0422} \\
\bottomrule
\end{tabular}
\centering
\label{tab::signal_dec_eg6}
\end{adjustwidth}
\end{table}

According to the results, we have the following conclusions:
\begin{itemize}
\item [(\romannumeral1)] The mode mixing problem is indeed a big challenge for some of the existing methods. For example, the maximum value of $x(t)$ is $2$, but the MAEs of the two components obtained by the SYNSQ\_STFT method are as high as $0.6620$ and $0.6778$, respectively, which basically does not separate the cosine signals with frequencies of 2.5Hz and 3.4Hz from $x(t)$.
\item [(\romannumeral2)] Many methods achieve satisfactory decomposition for $x(t)$. For example, it can be seen from the left plots in the 2nd and 3rd rows of Fig. \ref{fig:signal_dec_all} (a) that the components obtained by the EMD, INCMD, and RRCNN methods have relatively consistent oscillation modes with the ground truths. This viewpoint can also be drawn from Fig. \ref{fig:signal_dec_all} (b), although there are some obvious fluctuations, the TFE representations of the two components, obtained by EMD, INCMD, and RRCNN methods, are basically separated just like the those of the real components.
\item [(\romannumeral3)] Nonetheless, a closer look at the right plots in the 2nd and 3rd rows of Fig. \ref{fig:signal_dec_all} (a) reveals the subtle advantage of the RRCNN model at the boundaries. Due to the incompleteness of the waveform at the boundary, many existing methods are deeply affected by it, as are EMD and INCMD. However, the weights of the convolution filters in the RRCNN model depend on the entire waveform of the whole training samples, which reduce consistently the boundary effects.
\end{itemize}

\subsection{Can RRCNN be effective on noisy signals?}
\label{subsec:4}

Similar to the RRCNN inner loop block, we verify the robustness of RRCNN against additive Gaussian noise in this part. The constructed inputs and labels are listed in Tab. \ref{tab::inputs4}, where the inputs are generated by introducing additive Gaussian noise with the SNR set to $25dB$ to the signals in Tab. \ref{tab::inputs3}. After the RRCNN model is trained, we choose the signal consisting of two mono-components and additive Gaussian noise with a SNR of $15dB$ as the test data, which is given in Example 7. Since the smaller SNR value, the greater the noise, the noise of $x(t)$ is larger than that in the training data.

{\bf Example 7}: $x(t)=\cos(5\pi t)+\sin(8\pi t+2t^2+\cos(t))+\varepsilon(t)$, $t\in[0,6]$, SNR=$15dB$.

The errors between the ground truths and the components obtained by different methods, measured by MAE and RMSE, are reported in Tab. \ref{tab::signal_dec_noise_eg2}. Furthermore, the components, errors, and TFE representations of the three best performing methods, i.e., FDM, DCT\_GAS\_FDM, and RRCNN, are shown in Fig. \ref{fig:signal_dec_all} (d), (e), respectively.

According to the results, we can find that RRCNN works for the signals with additive Gaussian noise, although there is no overwhelming advantage, especially over FDM. Specifically, from the left plots in the 2nd-3rd rows of Fig. \ref{fig:signal_dec_all} (d), RRCNN basically separates the two mono-components, and the resulting components are consistent with the ground truths in the oscillation mode. Moreover, as shown in the right plots in the 2nd-3rd rows of Fig. \ref{fig:signal_dec_all} (d), the errors of RRCNN are relatively evenly dispersed in the entire time period, while those of the FDM and DCT\_GAS\_FDM methods are both small in the middle and large at the boundaries, which is consistent with the observations in Section \ref{subsec:3}.

\begin{table}[H]\scriptsize
\begin{adjustwidth}{0in}{0in}
\centering
\caption{Inputs disturbed by the Gaussian noise with the SNR set to $25dB$, and the labels used in Section \ref{subsec:4}, where $t\in[0,6]$.}
\renewcommand\arraystretch{1.2}
\begin{tabular}{ccccc}
\toprule
$x_1(t)$&  $x_2(t)$ & Inputs & Labels & Notes\\
\midrule
\multirow{2}{*}{$\cos(k\pi t)$} & $\cos((k+1.5)\pi t)$ &  \multirow{4}{*}{$x_1(t)+x_2(t)+\varepsilon(t)$} & \multirow{4}{*}{$[x_2(t),x_1(t)]$}  & \multirow{4}{*}{$k=5,6,\ldots,14$}  \\
& $\cos((k+1.5)\pi t+t^2+\cos(t))$ &  &  &    \\
\multirow{2}{*}{$0$} & $\cos((k+1.5)\pi t)$ & & &  \\
& $\cos((k+1.5)\pi t+t^2+\cos(t))$& & &  \\
\midrule
\multirow{2}{*}{$\cos(k\pi t)$} & $\cos(kl\pi t)$ & \multirow{4}{*}{$x_1(t)+x_2(t)+\varepsilon(t)$} & \multirow{4}{*}{$[x_2(t),x_1(t)]$}  &  \multirow{2}{*}{$k=5,6,\ldots,14$}  \\
& $\cos(kl\pi t+t^2+\cos(t))$ &  & &    \\
\multirow{2}{*}{$0$} & $\cos(kl\pi t)$ &  & &   \multirow{2}{*}{$l=2,3,\ldots, 19$}  \\
& $\cos(kl\pi t+t^2+\cos(t))$ & &&    \\
\bottomrule
\end{tabular}
\centering
\label{tab::inputs4}
\end{adjustwidth}
\end{table}

\begin{table}[H]\scriptsize
\begin{adjustwidth}{0in}{0in}
\centering
\caption{Metrics of the errors between the results obtained by different methods and the ground truths of Example 7.}
\renewcommand\arraystretch{1.2}
\begin{tabular}{ccccc ccccc}
\toprule
\multirow{2}{*}{Method} &  \multicolumn{2}{c}{$c_1$} & \multicolumn{2}{c}{$c_2$} & \multirow{2}{*}{Method} &  \multicolumn{2}{c}{$c_1$} & \multicolumn{2}{c}{$c_2$} \\
\cline{2-5} \cline{7-10}
& MAE & RMSE &  MAE & RMSE & & MAE & RMSE &  MAE & RMSE \\
\midrule
EMD\cite{huang1998empirical} & 0.1693 & 0.2961 & 0.2029 & 0.3485 & INCMD\cite{tu2020iterative} & 0.2436 &0.3870 &0.3343 &0.4720 \\
VMD\cite{dragomiretskiy2013variational} &0.1440 &0.2371 &0.1363 &0.2286 & SYNSQ\_CWT\cite{daubechies2011synchrosqueezed} & 0.5817& 0.6477& 0.0915 &0.1522 \\
EWT\cite{gilles2013empirical} &0.1840 &0.2945 &0.1754 & 0.2884 & SYNSQ\_STFT\cite{daubechies2011synchrosqueezed} &0.4353 &0.4992 &0.2009 &0.2463 \\
FDM\cite{singh2017fourier} & \textbf{0.1039} & \textbf{0.1765} & 0.0857 &0.1327  & DCT\_GAS\_FDM\cite{singh2018novel} &0.1288 &0.1801 & 0.0826& 0.1252\\
IF\cite{cicone2016adaptive}&0.1559  & 0.2318 &0.1601 & 0.2256  & EEMD\cite{wu2009ensemble} & 0.1870 & 0.2308 &0.1243 & 0.1925\\
 \textbf{RRCNN}& 0.1556 & 0.1977 & \textbf{0.0805} & \textbf{0.1014} \\
\bottomrule
\end{tabular}
\centering
\label{tab::signal_dec_noise_eg2}
\end{adjustwidth}
\end{table}

At last,  since RRCNN is a method designed with the help of CNN in the time domain, it can obtain an effect comparable to the existing methods in the time domain. Due to the lack of a prior information of RRCNN in the frequency domain, the effect of this method might be slightly reduced from the time-frequency domain. According to the results in Fig. \ref{fig:signal_dec_all} (e), the TFE distributions of the two mono-components obtained by the FDM, DCT\_GAS\_FDM, and RRCNN, are obviously spaced apart, but the TFE distribution of the component $c_1$ obtained by RRCNN has a much more severe jitter than the true component $\sin(8\pi t+2t^2+\cos(t))$ in the interval $t\in[2, 4]$. However, from the TFE representations, we can also see that the RRCNN method is able to reduce boundary effects compared to other methods.

\subsection{Can RRCNN be effective on the signals composed of orthogonal mono-components?}
\label{subsec:6}

In this part, we test the proposed RRCNN model on the signal composed of the orthogonal components. As discussed in Section \ref{subsec::RRCNN}, the RRCNN model in this part should have been equipped with the loss function with an orthogonal constraint, we directly add an inner product term to the loss function to control the orthogonality, i.e., $\gamma\frac{\vert<c_1, c_2>\vert}{\|c_1\|_2\|c_2\|_2}$, where $\gamma$ denotes the positive penalty parameter, and $c_1$ and $c_2$ are two components that are orthogonal. Then we train the model by the back propagation method based on the loss function. Although there is no guarantee of convergence in this case, it is simple, computationally efficient, and basically meets the expectation of orthogonality from the experimental results.

\begin{table}[H]\scriptsize
\begin{adjustwidth}{0in}{0in}
\centering
\caption{Inputs and labels used in Section \ref{subsec:6}, where $t\in[0, 2\pi]$.}
\renewcommand\arraystretch{1.2}
\begin{tabular}{ccccc}
\toprule
$x_1(t)$ &  $x_2(t)$ & Inputs & Labels & Notes\\
\midrule
\multirow{2}{*}{$\cos kt$} & \multirow{2}{*}{$\sin(k+l)t$} &\multirow{2}{*}{$x_1(t)+x_2(t)$} & \multirow{2}{*}{$[x_2(t), x_1(t)]$} & $k=6,7,8,9$ \\
 & & &  & $l=3,5,\ldots,33$ \\
\bottomrule
\end{tabular}
\centering
\label{tab::inputs6}
\end{adjustwidth}
\end{table}

The Fourier basis is adopted for the construction of the inputs of RRCNN because they are orthogonal. The constructed inputs and labels are given in Tab. \ref{tab::inputs6}. After the RRCNN is trained, we use it to decompose the signal given in Example 8, which is composed of two mono-components that are orthogonal and close in frequency.

{\bf Example 8}: $x(t)=\cos(7t)+\sin(9t)$, $t\in[0, 2\pi]$.

\begin{table}[H]\scriptsize
\begin{adjustwidth}{0in}{0in}
\centering
\caption{Metrics of the orthogonality, and errors of the obtained components by different methods in Example 8.}
\renewcommand\arraystretch{1.2}
\begin{tabular}{cccccc cccccc}
\toprule
\multirow{2}{*}{Method} & \multirow{2}{*}{$\rho(c_1, c_2)$} &  \multicolumn{2}{c}{$c_1$} &  \multicolumn{2}{c}{$c_2$}  & \multirow{2}{*}{Method} & \multirow{2}{*}{$\rho(c_1, c_2)$} &  \multicolumn{2}{c}{$c_1$} &  \multicolumn{2}{c}{$c_2$} \\
\cline{3-6} \cline{9-12}
& & MAE & RMSE &  MAE & RMSE & & & MAE & RMSE &  MAE & RMSE \\
\midrule
EMD\cite{huang1998empirical}& 0.1141 & 0.1796& 0.2776 & 0.2615 & 0.3581  & 
INCMD\cite{tu2020iterative} & 0.1785&0.1846 & 0.2485 &0.6205 & 0.7004 \\
VMD\cite{dragomiretskiy2013variational}& 0.4351& 0.5682 &0.6303 & 0.5682 &0.6303 & SYNSQ\_CWT\cite{daubechies2011synchrosqueezed}  &0.6268 &0.7697 & 0.9299 &0.4094 & 0.4823 \\
EWT\cite{gilles2013empirical} &\textbf{0.0000} & 0.6364 &0.7070 &0.6364 &0.7070 & SYNSQ\_STFT\cite{daubechies2011synchrosqueezed}  &0.1561 &0.3225 &0.3880 &0.3202 & 0.3835\\
IF\cite{cicone2016adaptive} & 0.6896  &0.2262 & 0.2776 & 0.2519 & 0.3092  & M-LFBF\cite{pustelnik2014empirical} &0.9543 &0.3858 &0.4721 &0.3857 & 0.4720  \\
FDM\cite{singh2017fourier} &0.0939 &0.1229 &0.2047 & 0.1244 & 0.2035 & DCT\_GAS\_FDM\cite{singh2018novel} &\textbf{0.0000} & 0.1222 & 0.1898 & 0.1211& 0.1824  \\
 \textbf{RRCNN}& 0.0595 &\textbf{0.1195} &\textbf{0.1759} & \textbf{0.0639} &\textbf{0.0889} \\
\bottomrule
\end{tabular}
\centering
\label{tab::signal_dec_orthogonal_eg1}
\end{adjustwidth}
\end{table}

The orthogonality and the errors between the resulting components and the corresponding ground truths are reported in Tab. \ref{tab::signal_dec_orthogonal_eg1}. According to the results, the EWT and DCT\_GAS\_FDM methods perform best in terms of orthogonality, which is attributed to the fact that the former is based on the wavelet transform and the latter is based on the discrete cosine transform, both of which have strong orthogonal constraint on the decomposition results. For RRCNN, its orthogonality is promoted by minimizing the loss function. Therefore, on one hand, the results of RRCNN tend to find a balance in each item of the loss function. On the other hand, the limited iterative process cannot ensure that the results are completely converges to the true solution. Combined with orthogonality and error metrics, the overall performance of the RRCNN model is still satisfactory. Specifically, it is not the best in terms of orthogonality, but it is also very close to orthogonality, outperforming the other models except EWT and DCT\_GAS\_FDM; in terms of error, its two components are also the closest to the true components.

Moreover, we select the top three methods in terms of the error metrics from Tab. \ref{tab::signal_dec_orthogonal_eg1}, that is, FDM, DCT\_GAS\_FDM, and RRCNN, and plot their obtained components, errors, and TFE representations in Fig. \ref{fig:signal_dec_all} (c), (f), respectively. From the plots, we can draw a conclusion similar to Example 6, that is, the RRCNN model can obtain the performance that is comparable to the state-of-the-art methods in the time domain, especially at the boundary, which can weaken the the impact of incomplete waveforms at the boundary to a certain extent. However, due to the lack of a priori information in the frequency domain in the design of RRCNN, its advantage in terms of TFE distribution might be reduced when compared with other methods especially in the middle of the signal. However its performance are overall better than existing methods at the boundaries, also in the time-frequency domain.

\subsection{Can RRCNN be effective on solutions of differential equations?}
\label{subsec:5}
To address this question, and following what has been done in the seminal work by Huang et al. \cite{huang1998empirical}, we test RRCNN against the solutions of a Duffing and Lorenz equations.
The labels are necessary in the training process of the RRCNN model, however, they are not known in this instance. Since the EMD method works well for  these two types of signals, we decided to use the results of EMD as the labels to verify the learning and generalization capabilities of the RRCNN model.

{\bf Example 9}: We consider the following Duffing equation: $\ddot{x}+\alpha x+\beta x^3=\gamma\cos(\omega t).$ We focus our attention on the decomposition of $\dot{x}(t)$. We first construct the inputs by solving the equation using the Euler method with the parameters set to $\alpha\in [-1, 0)$, $\beta\in[0.8, 1.2]$, $\omega\in[1.0, 1.4]$ with step size $0.1$, and $\gamma\in[0.04, 0.06]$ with step size $0.005$, respectively, where $t\in[0, 300]$ and the initial conditions: $\{x(0)=1, \dot{x}(0)=1\}$. And then, the first two IMFs of each input obtained by the EMD are collected as the labels.

\begin{figure}[!t]
\centering
%\begin{minipage}[t]{0.42\textwidth}\centering
\includegraphics[scale=0.45]{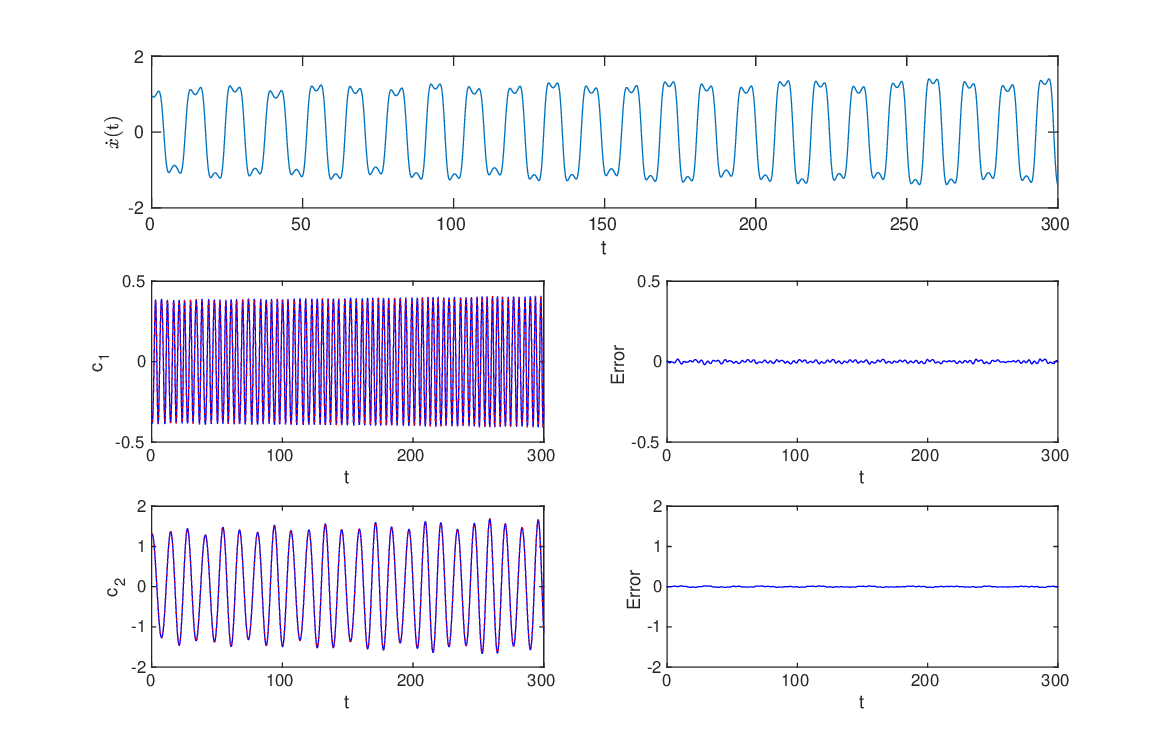}
\caption{Components of $\dot{x}(t)$ of Duffing equation with $\alpha=-0.85, \beta=1.05, \gamma=0.045, w=1.3$ by EMD and RRCNN. First row: the original signal. The 2nd-3rd rows, left: component obtained by EMD (blue solid curve) and RRCNN (red dotted curve); right: the error between the obtained components by EMD and RRCNN.}
\label{fig:signal_dec_duffing_eg1}
%\end{minipage}
\end{figure}

After the RRCNN is trained, we apply it to decompose the signal $\dot{x}(t)$ with $\alpha=-0.85, \beta=1.05, \gamma=0.045, w=1.3$, which was not included in the training data. The results are depicted in Fig. \ref{fig:signal_dec_duffing_eg1}. Both the resulting components produced by RRCNN are very close to those of EMD.

\begin{figure}[!t]
\centering
%\begin{minipage}[t]{0.42\textwidth}\centering
\includegraphics[scale=0.45]{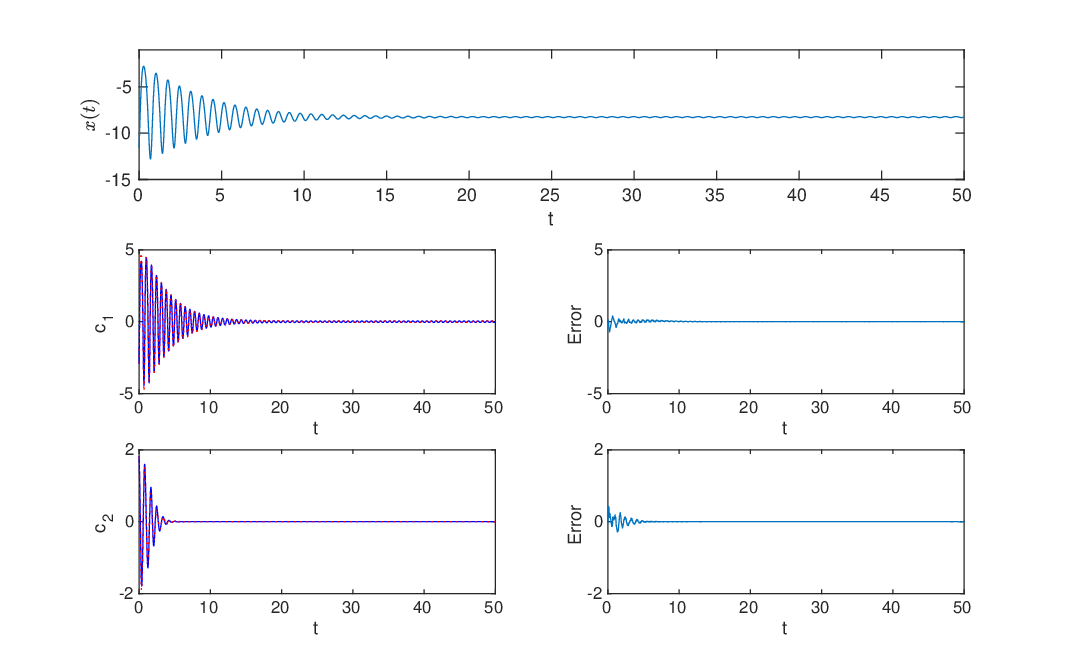}
\caption{Components of $x(t)$ of Lorenz equation with $\sigma=10.5, r=20.5, b=3.5$ by EMD and RRCNN. Same as Fig. \ref{fig:signal_dec_duffing_eg1}.}
\label{fig:signal_dec_lorenz_eg1}
%\end{minipage}
\end{figure}

{\bf Example 10}: The Lorenz equation is mathematically expressed as: $\dot{x}=-\sigma(x-y), \dot{y}=rx-y-xz, \dot{z}=-bz+xy.$ We take into account the decomposition of $x(t)$.
The inputs are the $x(t)$ achieved by the ode45 (code: https://www.mathwo rks.com/help/matlab/ref/ode45.html) method by changing the parameters $\sigma\in[9, 11]$ with step size $0.2$, $r\in[19, 21]$ with step size $0.2$, and $b\in[2,5]$ with step size $1$, where $t\in[0, 50]$ and initial conditions: $\{x(0)=-10, y(0)=0, z(0)=0\}$. Similarly to Example 10, we treat the first two IMFs of each input produced by the EMD as the labels. The results for signal $x(t)$ with parameters set to $\sigma=10.5, r=20.5, b=3.5$, predicted by the trained RRCNN model are shown in Fig. \ref{fig:signal_dec_lorenz_eg1}. The results show that the RRCNN has good learning and generalization capabilities for the solution to Lorenz equation, and can basically achieve the performance of EMD.

\subsection{Is RRCNN capable of processing real signals?}
\label{subsec:7}

We hereby employ the RRCNN model to process the real data, i.e., the length of day (LOD, data source: http://hpiers.obspm.fr/eoppc/eop/eopc04/eopc04.62-now, start date: Jan. 1,1962, end date: May 24, 2022), which is widely used in the verification of the signal decomposition methods.

\begin{figure}[H]
\centering
\subfigure[Example 11]{
\begin{minipage}[t]{0.48\linewidth}%\centering
\includegraphics[height=3.2in, width=3.6in]{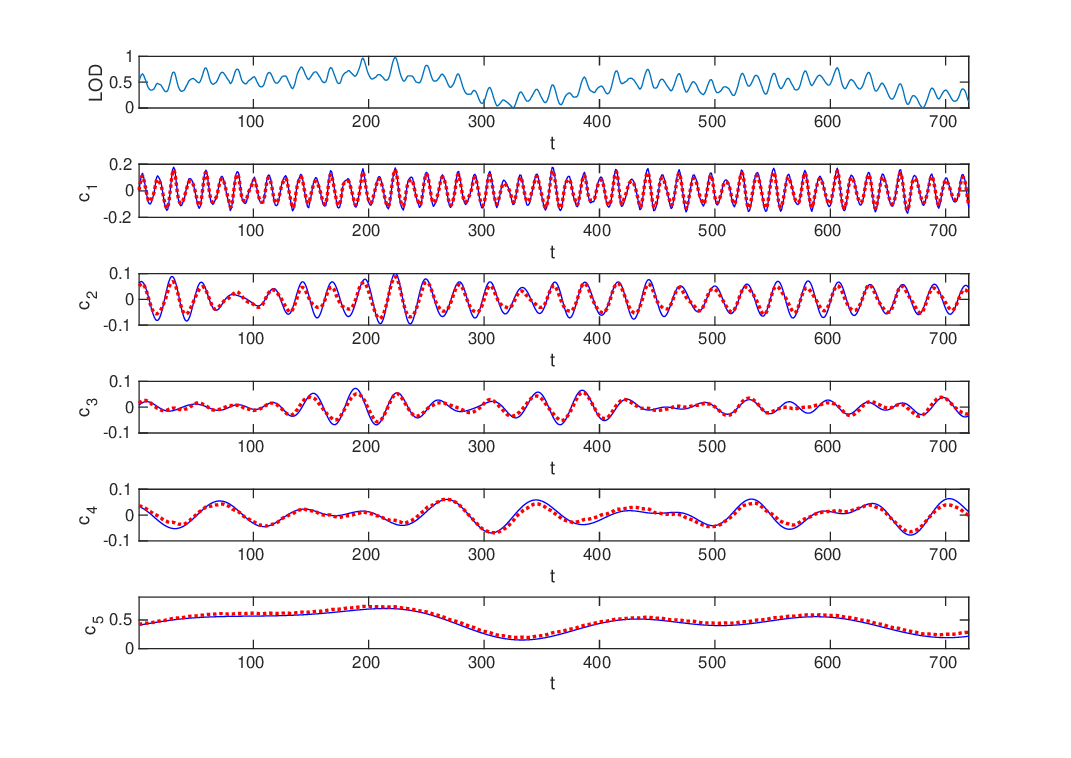}
\end{minipage}
}
\subfigure[Example 12]{
\begin{minipage}[t]{0.48\linewidth}%\centering
\includegraphics[height=3.2in, width=3.6in]{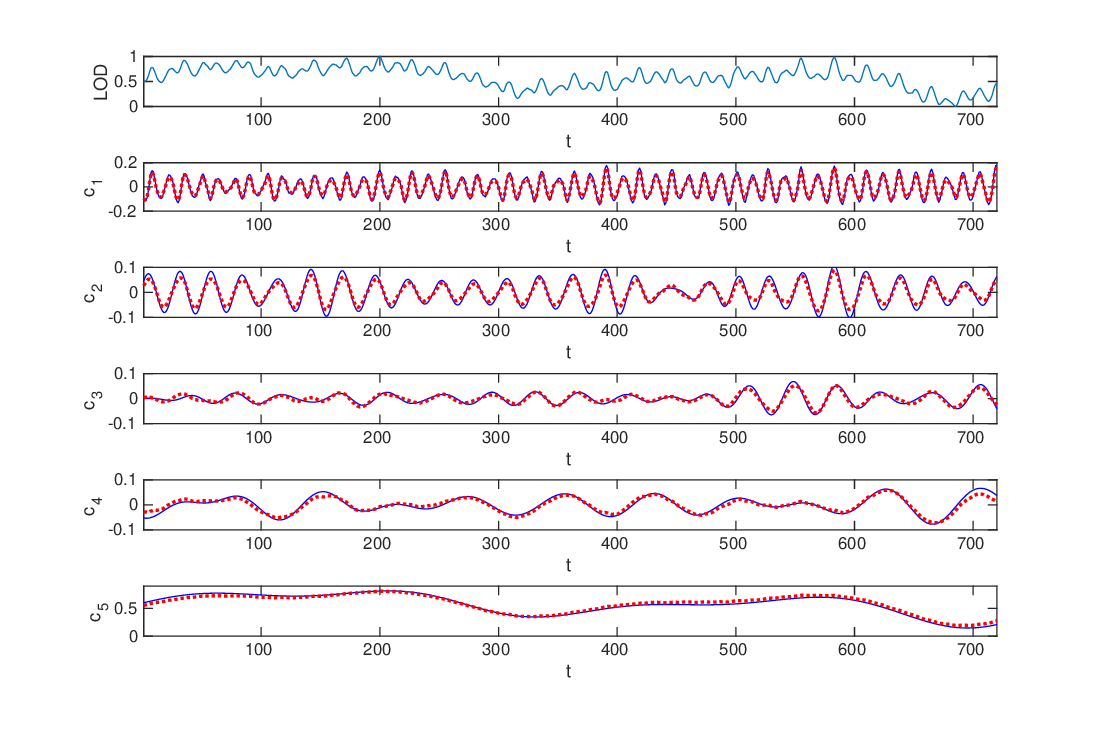}
\end{minipage}
}
\caption{Comparisons of the components obtained  by EWT and RRCNN for two LOD data given in Examples 11-12. Top panel: the original signal. The 2nd-6st panels: the obtained components by EWT (blue solid curve) and RRCNN (red dotted curve).}
\label{fig:signal_dec_lods}
\end{figure}

To generate the signals  from LOD data for use as the inputs to train the RRCNN model, we first split the LOD into a series of length $720$ (about two years) segments with a stride of $180$ (about half year). And then, similar to the situation in which we were dealing with the solutions of differential equations, the main challenge in training the RRCNN model on real signals is the ground truth calibration. We use the EWT method to produce the labels for each segment, because it can produce an artificially pre-set number of components.

{\bf Example 11}:  After the RRCNN model is trained, we apply it to the input, which is the LOD data ranging from Dec. 12, 2018 to Nov. 30, 2020, and is excluded in the training dataset.

{\bf Example 12}:  We also apply the trained RRCNN model to another input, which is the LOD data ranging from Dec. 7, 2019 to Nov. 25, 2021, and is excluded in the training dataset.

The components obtained from EWT and RRCNN for the input are depicted in Fig. \ref{fig:signal_dec_lods} (a)-(b). The plots prove that RRCNN can approximate EWT well.

\subsection{Computational time of different methods}
\label{sec::8}

To compare the computational time of the proposed method with other methods, we apply them to decompose the signal in Example 4, and record the corresponding runtime for comparison. Since in the prediction phase, RRCNN has the advantage of processing the decomposition of multiple signals in parallel, we also compare the running time of these methods when decomposing multiple signals. In this case, we no longer construct new signals to be decomposed, but just repeat the processing of the signal in Example 4, which does not affect the comparison of running time.

All the experiments are performed in Python 3.8.12 or Matlab R2020b on a Dell Precision 5820 tower with Intel(R) Xeon(R) W-2102 processor (2.90 GHz), 64G memory, and Ubuntu 18.04.3 operating system.

\begin{figure}[H]
\centering
\includegraphics[scale=0.45]{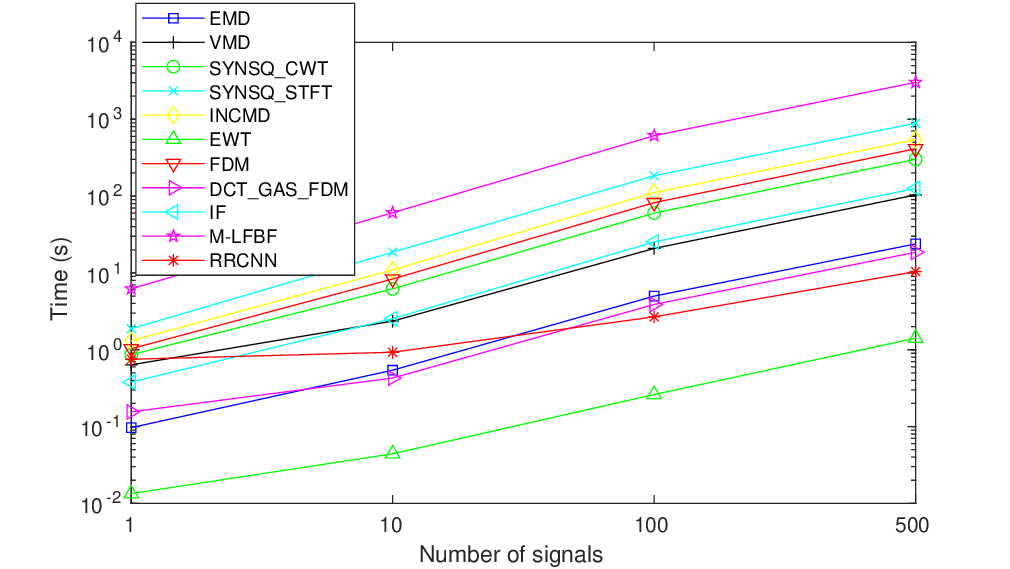}
\caption{Comparison of computational time of different signal decomposition methods when decomposing different numbers of signals.}
\label{fig:time_consuming}
\end{figure}

We depict the running time of different methods when decomposing different numbers of signals in Fig. \ref{fig:time_consuming}. From it, we have the following findings: When decomposing only one signal, the computational efficiency of RRCNN ranks in the middle among the $11$ methods compared in this work. However, as the number of signals increases to $10$, the computational efficiency of RRCNN begins to improve, and its ranking rises to 4th, surpassing the IF and VMD methods; when the number increases to 100, its computational efficiency ranks 2nd, beyond the EMD and DCT\_GAS\_FDM methods. Although there is still a gap between RRCNN and the EWT method that ranks 1st in computational efficiency, it can be seen that as the number of signals increases, the gap between them shrinks significantly.

\section{Conclusion}\label{sec:conclusion}
In the paper, we use deep learning techniques to tackle the issue of decomposing a non-stationary signal into oscillatory components. Specifically, we first construct the RRCNN inner loop block for obtaining the local average of a given signal, and then these blocks are cascaded into a deeper network, called RRCNN, which is used to decompose a given signal.

Since the proposed RRCNN model is based on the deep learning framework and is a supervised model, it has the advantages of these two types of models. First, the convolutional filter weights in the model are learnt according to the input signals, which makes the proposed method more adaptive. Second, some common tools in deep learning, like the residual structure and the nonlinear activation function, can be added to increase the expressive ability and flexibility of the proposed model. Third, the proposed model can be customized according to the application. For example, when processing signals composed of orthogonal components, an inner product term can be added to the loss function to enhance the orthogonality of the derived components. To verify the performance of the proposed model, we compare it with other existing models from seven aspects. And the artificial and real data are used in the experiments. All results show that the proposed method works better in handling the boundaries, mode mixing effects and the orthogonality of the decomposed components, and is more robust than the existing methods.

On the other hand, RRCNN has the classical limitations of supervised models. For example, the labels must be given in advance in the training phase. Therefore, the goal of this work is not to replace any existing method, but to propose a completely new kind of approach which is based on deep learning and supervised learning and to add it to the existing methods, to derive a more flexible and adaptive processing approach for signals. In the future, we plan to work on the extension of this work to multivariate signals. Furthermore, given that the main limitation of the RRCNN approach is that it requires other decomposition methods to calibrate the ground truths before the training phase, we plan to work in the future on the development of an unsupervised learning method to produce a new technique with similar performance to the RRCNN algorithm, but that does not require any training based on other techniques.

\section*{Declaration of Competing Interest}
No conflict of interest exits in the submission of this manuscript , and manuscript is approved by all authors for publication. We declare that the work described was original research that has not been published previously, and not under consideration for publication elsewhere, in whole or in part.

\section*{Acknowledge}
\label{sec::thanks}
F. Zhou research was partially supported by the National Natural Science Foundation of China [grant number 11901113], the Guangdong Basic and Applied Basic Research Foundation [grant number 2020A1515110951], and the China Scholarship Council [grant number 202008440024]. A. Cicone is a member of the Italian GNCS of the INdAM. H. Zhou research was partially supported by the National Science Foundation of US [grant number DMS-1830225], and the Office of Naval Research [grant number N00014-18-1-2852].

\section*{References}
%\bibliography{multi_RRCNN}

\end{document}